\documentclass{article}

\newlength\myboxwidth
\usepackage{booktabs}   
\usepackage{caption}    
\newcommand{\RNum}[1]{\uppercase\expandafter{\romannumeral #1\relax}}
\usepackage{colortbl}
\newcommand{\grayrow}{\rowcolor[gray]{.9}}
\usepackage{pifont}
\usepackage{multirow, adjustbox}

\usepackage[final]{neurips_2025}

\usepackage[utf8]{inputenc} 
\usepackage[T1]{fontenc}    
\usepackage[pagebackref,breaklinks,colorlinks]{hyperref}
\usepackage{url}            
\usepackage{booktabs}       
\usepackage{amsfonts}       
\usepackage{nicefrac}       
\usepackage{microtype}      
\usepackage{xcolor}         
\usepackage{xspace}
\usepackage{graphicx}
\usepackage{array}
\usepackage{pifont}
\usepackage{enumitem}
\usepackage{subcaption}
\usepackage{hyperref}
\usepackage{float}
\usepackage{bm}
\usepackage{comment}

\newcolumntype{R}[1]{>{\rotatebox[origin=c]{45}{#1}}c}
\newcommand{\cmark}{\textcolor{green!60!black}{\ding{51}}}

\def\projdataset{egoEMOTION\xspace}

\title{

\projdataset: Egocentric Vision and Physiological Signals for Emotion and Personality  Recognition in Real-World Tasks

}

\author{
  Matthias Jammot\thanks{Equal contribution}~, Björn Braun\footnotemark[1]~, Paul Streli, Rafael Wampfler, Christian Holz \\
  Department of Computer Science\\
  ETH Zurich, Switzerland \\[.5em]
  {\small\href{https://siplab.org/projects/egoEMOTION}{\color{magenta}{\texttt{https://siplab.org/projects/egoEMOTION}}}}%
  {}
  \vspace{-5mm}
}

\begin{document}

\maketitle

\begin{abstract}

Understanding affect is central to anticipating human behavior, yet current egocentric vision benchmarks largely ignore the person's emotional states that shape their decisions and actions.
Existing tasks in egocentric perception focus on physical activities, hand-object interactions, and attention modeling---assuming neutral affect and uniform personality.
This limits the ability of vision systems to capture key internal drivers of behavior.
In this paper, we present \emph{\projdataset}, the first dataset that couples egocentric visual and physiological signals with dense self-reports of emotion and personality across controlled and real-world scenarios.
Our dataset includes over 50~hours of recordings from 43~participants, captured using Meta's Project Aria glasses.
Each session provides synchronized eye-tracking video, head-mounted photoplethysmography, inertial motion data, and physiological baselines for reference.
Participants completed emotion-elicitation tasks and naturalistic activities while self-reporting their affective state using the Circumplex Model and Mikels' Wheel as well as their personality via the Big Five model.
We define three benchmark tasks:
(1)~continuous affect classification (valence, arousal, dominance);
(2)~discrete emotion classification; and
(3)~trait-level personality inference.
We show that a classical learning-based method, as a simple baseline in real-world affect prediction, produces better estimates from signals captured on egocentric vision systems than processing physiological signals.
Our dataset establishes emotion and personality as core dimensions in egocentric perception and opens new directions in affect-driven modeling of behavior, intent, and interaction.

\end{abstract}

\section{Introduction}
\label{sec:introduction}

Egocentric vision systems are well positioned to capture the signals for modeling human attention, interaction, and behavior in real-world environments.
Benchmarks in this area have driven advances in action recognition~\cite{garcia2018first}, object manipulation~\cite{kwon2021h2o, ohkawa2023assemblyhands}, gaze prediction~\cite{kellnhofer2019gaze360}, and interaction understanding~\cite{ego4d, egoexo4d}.
These tasks focus on what people do and attend to, using first-person visual input to model external behavior~\cite{ ego4d, egoexo4d, Nymeria}.
Such progress has expanded the scope of perception systems, in domains such as Mixed Reality~\cite{visionpro, quest3}, front-line productivity work~\cite{cheng2023mixedreality}, and context-aware interaction~\cite{braun2025egoppg, hartig2025multimodal}.
However, current benchmarks overlook internal states like emotion and personality that shape these behaviors, implicitly assuming affect-neutral and behaviorally uniform participants~\cite{egoexo4d}, ignoring individual differences.
This limits how egocentric systems can model behavior that depends on mood, arousal, or personality traits~\cite{ego4d}.
Tasks involving decisions~\cite{mozikov2024eai}, social interaction~\cite{ego4d}, and memory~\cite{ochsner2000social} require grounding in affect.
We argue that without such affective modeling, emerging egocentric platforms cannot fully understand human behavior.

In this paper, we present \projdataset, a dataset for affect and personality recognition from egocentric visual and physiological signals.
Our dataset thus addresses the current gap in egocentric vision by supplying emotional and trait labels grounded in self-reports.
\projdataset comprises synchronized multimodal data during both emotion-elicitation protocols and naturalistic tasks, such as watching video clips, painting, playing social and video games.
Each session captures a participant's eye-tracking video, inertial motion (IMU), outward point-of-view (POV) camera, and photoplethysmogram (PPG) to gauge cardiac activity from Meta's Project Aria glasses~\cite{ProjectAria}, as well as physiological baseline measurements, including electrocardiograms (ECG), respiratory rates (RSP), and electrodermal activity (EDA)---all suitable to extract indicators of a person's affective state~\cite{AMIGOS, ASCERTAIN}.
Participants reported their affect using the Circumplex Model~\cite{russellcircumplex} and Mikels' Wheel~\cite{mikelswheel} and assessed their personality using the Big Five model~\cite{costa_bigfive}.
In total, our dataset spans 50~hours of recordings from 43~participants across varied emotional and social contexts.

We then define three prediction benchmarks---continuous affect regression, discrete emotion classification, and personality inference--- and provide baselines showing that egocentric signals, particularly eye-tracking features, outperform traditional physiological baselines in real-world emotion prediction.
This highlights the promise of affective modeling from egocentric vision systems and establishes \projdataset as a foundation for future research in this direction.


Collectively, we contribute:

\begin{enumerate}[leftmargin=*]
    \item the first multimodal dataset that uses an egocentric vision system for emotion and personality recognition. Our dataset comprises both induced and naturalistic tasks that cover a wide range of elicited emotions, while offering nuanced mixed-emotions self-reporting.
    
    \item three benchmark tasks and associated baseline: valence-arousal-dominance, discrete emotion, and personality recognition.
    Our results show that using features solely from egocentric vision systems outperforms estimates from physiological signals.
    
    \item an open-source release of our ethics-approved dataset and baseline implementations (\texttt{23 ETHICS-008}).

\end{enumerate}

\begin{figure}
  \centering
  \includegraphics[width=\textwidth]{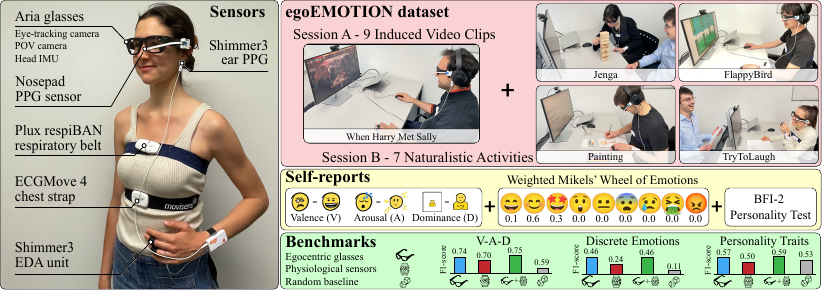}
  \caption{\textbf{\projdataset is a multimodal emotion and personality recognition dataset that captures participants' facial, eye-tracking, egocentric, and physiological signals during induced video stimuli and naturalistic real-world activities}. Participants reported their emotions via emoti-SAM~\cite{emoti-sam} and a weighted Mikels' Wheel~\cite{mikelswheel}, and their personality using the Big Five model~\cite{costa_bigfive}.}
  \label{fig:PEPPER_overview}
\vspace{-1em}
\end{figure}
\section{Related Work}
\label{sec:related_work}

\addtolength{\tabcolsep}{-0.15em}
\begin{table}
    \centering
    \caption{\textbf{Comparison of public multimodal affective datasets.}}
    \small

    \begin{tabular}{p{4cm} r p{0.0cm} c  *{5}{p{0.0cm}} c  *{10}{p{0.0cm}} c *{4}{p{0.0cm}}}
    \toprule[1.5pt]

        \multicolumn{4}{l}{} & 
        \multicolumn{5}{c}{\textbf{Elicitation}} & 
         &
        \multicolumn{11}{c}{\textbf{Sensing Modalities}} & 
        
        \multicolumn{4}{c}{\textbf{Annotation}} \\
        \cmidrule(lr){5-9}
        \cmidrule(lr){11-21}
        \cmidrule(lr){22-25}
    
         \textbf{Dataset (year)} & 
         \rotatebox[origin=l]{90}{No. Subjects} &  
         \rotatebox[origin=l]{90}{Mobile Sensors} & 
          &
         \rotatebox[origin=l]{90}{Induced} & 
         \rotatebox[origin=l]{90}{Natural} & 
          &
         \rotatebox[origin=l]{90}{Individual} & 
         \rotatebox[origin=l]{90}{Social} & 
          &
         \rotatebox[origin=l]{90}{BVP} & 
         \rotatebox[origin=l]{90}{ECG} & 
         \rotatebox[origin=l]{90}{EDA} & 
         \rotatebox[origin=l]{90}{EEG} & 
         \rotatebox[origin=l]{90}{Eye Tracking} &  
         \rotatebox[origin=l]{90}{Face} & 
         \rotatebox[origin=l]{90}{IMU (head)} & 
         \rotatebox[origin=l]{90}{IMU (wrist)} & 
         \rotatebox[origin=l]{90}{POV camera} & 
         \rotatebox[origin=l]{90}{RSP} & 
          &
         \rotatebox[origin=l]{90}{A-V (-D)} & 
         \rotatebox[origin=l]{90}{Emotion Tags} & 
         \rotatebox[origin=l]{90}{Weighted Tags} & 
         \rotatebox[origin=l]{90}{Big-5}
            
         \\
        \midrule
        \midrule
        \noalign{\vskip 3pt}
        \textbf{\projdataset (2025)} & 43 & \cmark & & \cmark & \cmark & & \cmark & \cmark & & \cmark & \cmark & \cmark &  & \cmark & \cmark & \cmark & \cmark & \cmark & \cmark & & \cmark & \cmark & \cmark & \cmark \\
        \noalign{\vskip 3pt}
        EmoPairCompete~\cite{das2024emopaircompete} (2024) & 28 & \cmark & &  & \cmark & &  & \cmark & &  \cmark &  & \cmark &  &  &  &  & \cmark &  &  & &  & \cmark & \cmark &  \\
        G-REx \cite{G-REx} (2024) & 191 & \cmark & &  & \cmark & &  & \cmark & & \cmark &  & \cmark &  &  &  &  &  &  &  & & \cmark &  &  &  \\
        eSEE-d \cite{skaramagkas2023esee} (2023) & 48 &  && \cmark &  && \cmark &  &&  &  &  &  & \cmark &  &  &  &  &  & &  & \cmark & \cmark &  \\
        BIRAFFE2 \cite{biraffe2} (2022) & 102 & \cmark &&  & \cmark && \cmark &  &&   & \cmark & \cmark &  &  &  &  & \cmark &  &  & & \cmark &  &  & \cmark \\
        PPB-Emo \cite{PPB-Emo} (2022) & 40 &  &&  & \cmark && \cmark &  &&  & \cmark &  & \cmark &  & \cmark &  &  &  &  && \cmark & \cmark &  &  \\ 
        K-EMOCON \cite{K-EMOCON} (2020) & 21 & \cmark & &  & \cmark & &  & \cmark & & \cmark & \cmark & \cmark & \cmark &   & \cmark &   & \cmark &  &   & & \cmark & \cmark &  & \\
        AMIGOS \cite{AMIGOS} (2018) & 40 & \cmark && \cmark &  && \cmark & \cmark &&  & \cmark & \cmark &  \cmark &  & \cmark &  &   &   &   && \cmark & \cmark &  & \cmark \\
        ASCERTAIN \cite{ASCERTAIN} (2016) & 58 & \cmark & & \cmark &  & & \cmark &  & &   & \cmark & \cmark & \cmark &   & \cmark &  &   &   &   & & \cmark &  &  & \cmark \\
        DEAP \cite{DEAP} (2012) &32 &  & & \cmark &  & & \cmark &  & & \cmark &  & \cmark & \cmark &  &   &  &  &  & \cmark & & \cmark &  &  &  \\
        MAHNOB-HCI \cite{MAHNHOB-HCI} (2012) & 27 &  & & \cmark &  & & \cmark &  & &  & \cmark & \cmark & \cmark & \cmark & \cmark  &  & \cmark &   & \cmark & & \cmark & \cmark &  &  \\

    \bottomrule[1.5pt]
    \multicolumn{25}{l}{\small{Datasets where participants were shown videos are classified as `induced' elicitation.}} \\
    \multicolumn{25}{l}{\small{A: Arousal, V: Valence, D: Dominance. BVP: Blood Volume Pressure (from PPG sensor).}}
    \end{tabular}
    \label{tab:affect_datasets_physiological}
\vspace{-1em}
\end{table}
\addtolength{\tabcolsep}{0.15em}

Emotion elicitation can be either induced, using predefined stimuli like videos or sounds, or naturalistic, arising spontaneously in real-life contexts. The terms \textit{in-the-wild}, \textit{real-world}, or \textit{naturalistic} data have been denoted to describe data collection when the experimenters do not control the emotion elicitation nor constrain the data acquisition~\cite{larradet2020toward}. These emotional responses may occur in either static environments, where participants remain still (workplace, car, cinema), or ambulatory environments, where data is collected during everyday activities~\cite{larradet2020toward}.

\paragraph{Induced.} Due to the challenges of collecting physiological data in real-world settings~\cite{moebus2024nightbeat,picard1997affective, smets2019into}, many emotion recognition studies have been conducted in controlled laboratory settings and have used pre-selected video clips as emotional stimuli, as shown in Table \ref{tab:affect_datasets_physiological}. DEAP~\cite{DEAP} collected electroencephalogram (EEG), facial, and physiological data from 32 participants who self-reported their emotions using valence-arousal-dominance (V-A-D) ratings after viewing 40 1-minute-long music videos. MAHNHOB-HCI~\cite{MAHNHOB-HCI} collected signals similar to those of DEAP with the addition of audio and eye gaze data. Their study gathered 27 participants who, after watching 20 short videos in a first experiment, followed by 28 images and 14 videos in a second experiment, annotated their emotions using V-A-D rating scales and emotional tags.  
ASCERTAIN~\cite{ASCERTAIN} extended these studies by using wireless physiological sensors and facial features from 58 participants, while also capturing personality traits through the Big Five model~\cite{costa_bigfive}. AMIGOS~\cite{AMIGOS} further advanced the field by introducing group-based video viewing and assessing mood in parallel to emotions and personality.

\paragraph{Naturalistic.} 
While controlled lab settings are useful for isolating variables and evaluating specific emotions, their ecological validity is limited, raising concerns about real-world applicability~\cite{larradet2020toward, DAPPER}. 
The G-REx~\cite{G-REx}, EmoPairCompete~\cite{das2024emopaircompete}, and K-EmoCon~\cite{K-EMOCON} datasets naturally induced emotions in participants through group movie sessions, solving puzzles in pairs, and paired debates, respectively. BIRAFFE2~\cite{biraffe2} exposed its participants to IAPS~\cite{lang1997iaps} visual and IADS~\cite{IADS} audio stimuli, followed by three mini-quest video games. PPB-Emo~\cite{PPB-Emo} recorded participants in a driving simulator. 
While these datasets have advanced emotion recognition in static real-world tasks, they remain limited in the array of sensors used and their range of emotionally-diverse activities. 
To vary naturalistic emotions recorded, emotion recognition has been investigated in ambulatory real-world settings, using mobile phones to self-report emotions~\cite{kovacevic, schmidt2019, DAPPER, wampfler2022} and personality~\cite{kovavcevic2023personality}.
However, these in-the-wild studies face key limitations: self-reports are often infrequent and intrusive, the lack of known stimuli hinders interpretation of physiological responses, and signal quality is affected by motion artifacts and inconsistent sensor use.

\paragraph{Egocentric.} 
The rise of mobile egocentric systems has enabled large-scale~\cite{glabella}, in-the-wild datasets such as EPIC-KITCHENS~\cite{EPICKITCHENS}, Ego4D~\cite{ego4d}, Ego-Exo4D~\cite{egoexo4d}, and Nymeria~\cite{Nymeria}, supporting tasks like activity recognition and social behavior modeling. 
However, these datasets assume neutral affect and lack emotional context, limiting their use for modeling user intent or emotion-driven behaviors. 
Integrating emotion recognition could enable more affect-aware activity analysis and adaptive human-AI interaction. 
In this context, the eye-tracking videos recorded with our glasses offer a valuable modality for capturing users’ intrinsic emotional states during diverse real-world tasks. 
MAHNOB-HCI~\cite{MAHNHOB-HCI} was one of the first datasets to introduce eye-tracking as a modality.
While emotion recognition using \textit{mobile} eye-tracking systems has been explored~\cite{kwon2021glasses, SPIDERS, tarnowski2020eye}, their datasets were not made public and gathered few participants.
To date, eSEE-d~\cite{skaramagkas2023esee} is the only public dataset for emotion recognition using mobile eye-tracking. 
However, its limited four-emotion questionnaire, absence of physiological signals, and controlled setup (e.g., chin rest) reduce its validity for real-world applications.
While heart rate can be estimated from facial videos~\cite{braun2024suboptimal, poh2010non, yu2022physformer}, other physiological signals, such as EDA, remain challenging to estimate~\cite{braun2023video, braun2024sympcam} but are significant for judging a person's emotional response~\cite{luong2022characterizing}.
Personality recognition from mobile eye-tracking systems has been explored by Hoppe et al.~\cite{hoppe2018eye} with participants walking on a university campus and Berkovsky et al.~\cite {personalityET2019}, in which participants watched images from the IAPS dataset~\cite{lang1997iaps} in laboratory settings. Neither of these studies recorded physiological signals, nor released their dataset.

\section{{\projdataset} Dataset}
\label{sec:dataset}

Prior work on emotion and personality recognition using physiological sensors has typically focused on affect, valence, and personality recognition---often in controlled lab settings with specialized equipment.
We go further by collecting detailed emotion self-reports alongside affect, valence, and personality, across both induced and naturalistic tasks (see Figure~\ref{fig:PEPPER_overview}). 
While our setup includes both standard physiological sensors and an egocentric vision system, we show that egocentric video alone is sufficient to enable practical, real-world applicability beyond traditional sensor-based approaches.

\subsection{Dataset Design}
\subsubsection{Experimental Protocol}

\addtolength{\tabcolsep}{-0.2em}
\begin{table}[t]
\centering
\caption{\textbf{Summary of emotional elicitation tasks}: induced (\texttt{1--9}) and naturalistic (\texttt{10--16}).}
\begin{tabular}{@{}p{1.8cm}p{0.45cm}p{9.5cm}p{1.3cm}@{}}
\toprule[1.5pt]
\textbf{Activity} & \textbf{ID} & \textbf{Description}  & \textbf{Duration}\\
\midrule
\midrule
\textit{Video Clips$^*$} & \texttt{1--9} &  AnimalCruelty, AuroraBorealis, BearGrylls, CollegeAcceptance, HarrySally, JoJoRabbit, LoveActually, MovingShapes, Psycho   & 9 $\times$ 48 s \\
\midrule
\textit{Flappy Bird} & \texttt{10} & Click to keep a bird flying through pipes. Restart upon failure. & 4 min \\
\textit{Jelly Bean} & \texttt{11} & Eat three unpleasant-tasting jelly beans. & 2 min \\
\textit{Jenga} & \texttt{12} & Remove blocks from a tower without collapsing it with experimenter. & 5 min \\
\textit{Painting} & \texttt{13} & Paint with brushes and crayons, listening to \textit{Your Song} (Elton John). & 4 min \\
\textit{Sad Letter} & \texttt{14} & Write a letter to someone lost, listening to \textit{Adagio for Strings}. & 4 min \\
\textit{Slenderman} & \texttt{15} & Find eight pages in dark woods while escaping the Slenderman. & 6 min \\
\textit{Try to Laugh} & \texttt{16} & Take turns with experimenter telling pre-written jokes. & 4 min \\
\bottomrule[1.5pt]
 \multicolumn{4}{l}{$^*$\small{A detailed description of the emotion-inducing video clips is presented Table~\ref{tab:description_video_clips} of Appendix \ref{app:study_protocol}.}}
\end{tabular}
\label{tab:naturalistic_activities}
\vspace{-1em}
\end{table}

Upon arrival, we explained the study protocol and self-report questionnaires to the participants, asked them to sign a consent form, and then equipped them with the sensors.
The experimental protocol (see Figure~\ref{fig:PEPPER_overview}) consisted of two sessions (A and B) with a total of 16 different tasks.
We conducted the experiment in a regular office next to a window, with the experimenter seated behind a curtain to avoid affecting participants' emotional reactions.
Before starting session A, each participant performed an eye-tracking calibration.
In session A, participants watched nine video clips ($\mu$ = 48~s, see Table~\ref{tab:naturalistic_activities}) corresponding to the eight emotions from Mikels' Wheel \cite{mikelswheel}, plus a ninth neutral emotion.
All videos were extensively validated by previous work to elicit target emotions~\cite{MAHNHOB-HCI, ASCERTAIN}.
Before each video clip, participants had to watch a 5-second video of a fixation cross to refocus their gaze~\cite{AMIGOS, ASCERTAIN}.
In session B, participants conducted seven activities (see Table~\ref{tab:naturalistic_activities}) that we selected to reflect spontaneous everyday activities to further the study's ecological validity \cite{rottenberg2007emotion}.
We designed the activities to minimize physical effort to avoid activity-induced variations in the recorded signals. 
After each task in sessions A and B, participants self-reported their perceived emotions.
They were instructed to report their true emotion, not the one they perceived as being the `correct' one.
The questionnaires completed, they watched a neutral video of clouds to mitigate any carry-over effect of the previous emotional stimulus.
The task order in both sessions was randomized for each participant.

\subsubsection{Data Annotation}

\begin{figure}
  \centering
  \includegraphics[width=\textwidth]{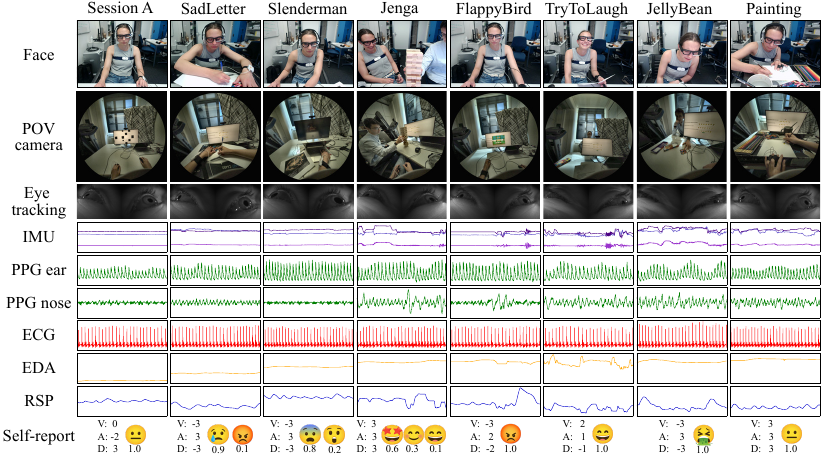}
  \caption{\textbf{Data collection} from the egocentric glasses and physiological sensors during each task with their associated self-reports. Further information about the study protocol is available in Appendix~\ref{app:study_protocol}.}
  \label{fig:PEPPER_sensors}
  \vspace{-1em}
\end{figure}

The experiment was performed on a graphical user interface coded using the \texttt{PyQt5} Python library, which would successively show washout, emotional stimulus, and self-report.
For session B, the experimenter would explain and setup the upcoming activity once the washout video was completed. 
Participants used their dominant hand to navigate through two questionnaires using a serial mouse. 
The first questionnaire, built on the Circumplex Model of Affect~\cite{russellcircumplex}, comprised a self-assessment manikin (SAM) where participants reported the arousal (A), valence (V), and dominance (D) of their emotion. 
We used the 7-point emoji-based emoti-SAM~\cite{emoti-sam}, which balanced response granularity with cognitive load and was intuitive given the ubiquity of emojis.
If SAM ratings are standard in emotion recognition studies, they are usually reduced to binary labels (e.g., \emph{high} / \emph{low}), which oversimplifies emotions.
As exemplified in Figure~\ref{fig:PEPPER_sensors}, the distinct emotions of fear, sadness, disgust, and anger all fall into the low-valence / high-arousal quadrant, making them difficult to distinguish.
Some studies have introduced binary emotional tags to address this~\cite{Gebhardt, AMIGOS, MAHNHOB-HCI}, but these lack nuance, particularly when mixed emotions are present, since each emotion carries equal weight in the analysis.

To gain nuance, we used as second questionnaire a weighted version of emotional tags. 
The participants distributed 100\% across nine emotions (eight from Mikels' wheel~\cite{mikelswheel} plus a neutral option) in increments of 10\%, ensuring the weights sum to unity for every report. The emotions were Amusement (Amu), Content (Con), Excitement (Exc), Awe, Fear (Fea), Sadness (Sad), Disgust (Dis), Anger (Ang), and Neutral (Neu).
This captured the relative emotional strength perceived by the user, distinguishing between stimuli that have one dominating emotion and others where emotions are more homogeneous. 
This annotation allowed complex emotions to be represented as vectors with attributes such as polarity, type, intensity, similarity, and additivity, following Yang et al.~\cite{yang2021circular}. 
It also enabled identifying the dominant emotion, leading to a more precise 9-class classification.

Finally, participants filled in the Big Five Inventory-2 (BFI-2) personality questionnaire online\,\cite{bfi2} before the experiment. The BFI-2 assesses five major personality traits: Extraversion (\emph{Ex}), Agreeableness (\emph{Ag}), Conscientiousness (\emph{Co}), Negative Emotionality (\emph{NE}), and Open-Mindedness (\emph{OM}).

\subsubsection{Sensors}
Our study used mobile wearable sensors to capture participants emotional responses, as shown in Figure~\ref{fig:PEPPER_overview}. 
We used the Project Aria glasses~\cite{ProjectAria} for their significant promise in capturing ecologically valid data that does not inhibit natural activities and behavior of the participants. 
Using the device's `Profile 16', we recorded eye-tracking (ET) videos with a 640 $\times$ 480 pixel resolution per eye at 90~fps, egocentric vision through a 1408 $\times$ 1408 POV RGB camera at 10~fps, and head movements through two IMUs sampling at 1000~Hz and 800~Hz. 
We supplemented the egocentric glasses with an in-house nosepad PPG sensor sampling at 128~Hz.
A Shimmer3 unit recorded PPG and EDA signals at the ear and fingers, respectively, at 256~Hz. 
A 1024~Hz Movisens ECG4Move4 chest belt measured the participant's ECG data while a plux respiBAN respiratory belt measured their respiration pattern~(RSP) at 400~Hz.
We recorded participants' facial expressions with a 60~fps 1280~$\times$~720 webcam for external labeling of emotions.

\subsection{Recruitment and Recording}
We recruited 43 healthy participants, mostly students, voluntarily with a CHF 30 compensation. 
The 24 female and 19 male participants were between 19 and 29 years old ($\mu$ = 26, $\sigma$ = 2). 
Based on the Fitzpatrick scale \cite{fitzpatrick}, 3 participants had skin type I, 19 had skin type II, 9 had skin type III, 8 had skin type IV, and 5 had skin type V. 
Each participant was recorded in a single session that lasted approximately 105 minutes. 
They had to confirm that they were not taking tranquilizers, psychotropic drugs, or narcotics, and were not diagnosed with any cardiovascular disease. 
They were also informed that they would have to carry two belts (ECG and RSP) on their chest. 
We ensured that all participants had sufficient English proficiency to understand the videos that they were shown.

\subsection{Dataset Composition}

Participant recordings were cut to the duration of session A ($\mu$ = 20 min) and session B ($\mu$ = 49 min). 
The dataset contains all raw sensor streams presented in Figure~\ref{fig:PEPPER_sensors}, each preserved at its sampling frequency. 
Since all sensors were equipped with IMUs, they were synchronized at the start and end of each experiment by simultaneously shaking them. This yields \projdataset, a dataset composed of 43 participant recordings, each completing 16 emotional tasks.
In total, the dataset offers over 50 hours of synchronized (90 Hz) multimodal data of egocentric and physiological signals. 
The dataset is structured by participant, with each folder containing the corresponding sensor streams. 
The start and end of each task were manually labelled to enable task-specific analyses.
\section{Dataset Descriptives}
\label{sec:data_analysis}

\subsection{Analysis of self-reports}

\begin{figure}[t]
    \includegraphics[width=\textwidth]{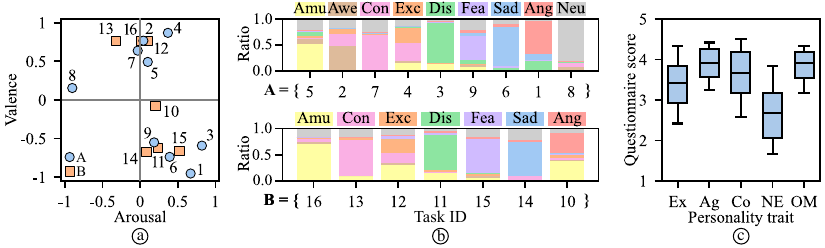}
    \caption{\textbf{Participant self-reports across tasks}. \textbf{(a)} Mean arousal-valence ratings. \textbf{(b)} Proportions of discrete emotions reported in Sessions A and B. \textbf{(c)} Boxplots of Big Five personality trait scores.}
    \label{fig:self_reports_all}
    \vspace{-1em}
\end{figure}

Figure~\ref{fig:self_reports_all} shows the self-reports across all participants of (a) mean arousal-valence ratings for each video clip and task, (b) discrete emotions (dominant emotion indicated above task), and (c) personality traits. 
While participants reported a wide range of valence, arousal ratings were less varied and consistently high across tasks.
The video clips and naturalistic activities elicited a diverse range of emotions (see Figure~\ref{fig:self_reports_all}b).
The naturalistic activities elicited stronger emotional responses, evidenced by a lower proportion of neutral tags.
More detailed information is provided in Appendix~\ref{app:additional_dataset_descriptives}. 

\subsection{Correlations}
\begin{figure}[t]
    \includegraphics[width=\textwidth]{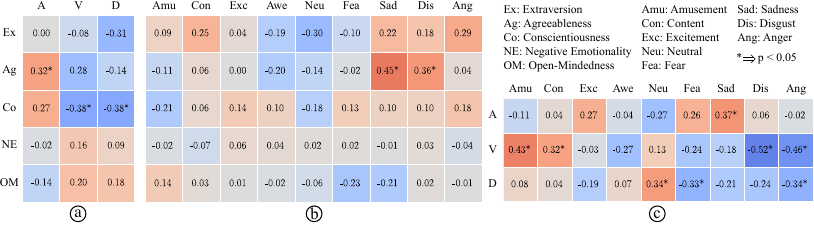}
    \caption{\textbf{Pearson correlations between self-reports.} \textbf{(a)} continuous self-ratings and personality scores \textbf{(b)} discrete emotions and personality scores \textbf{(c)} discrete emotions and continuous self-ratings.}
    \label{fig:correlation_continuous_emotion_personality_A_and_B}
    \vspace{-1em}
\end{figure}

Figure~\ref{fig:correlation_continuous_emotion_personality_A_and_B} presents the Pearson correlation matrices between continuous affect self-ratings (A-V-D), discrete emotions, and personality traits.
Focusing on significant correlations, \emph{Co} was negatively correlated with both V and D.
\emph{Ag} showed a moderate positive correlation with A.
The strongest relationships for discrete emotions and personality were between \emph{Ag} and Sadness and Disgust.
The correlations between discrete emotional states and the continuous affective dimensions were the following: V was positively associated with Amusement and Content, while negatively associated with Disgust and Anger, aligning with expectations. 
D showed significant negative correlations with Disgust and Anger, while A was positively linked to Excitement and Sadness.
Appendix~\ref{app:additional_dataset_descriptives} contains more details on the individual correlations between the self-reported annotations in Sessions A and B.

\section{Baselines}
\label{sec:baselines}

To demonstrate the benefits of \projdataset and motivate follow-up research, we propose three benchmarking tasks: predicting a participant's self-reported \emph{affective state}, \emph{discrete emotions}, and \emph{personality traits}.
We evaluate classical machine learning methods using data from wearable sensors (ECG, EDA, RSP) and the Aria glasses (accelerometers, eye-tracking, nosepad PPG). We contrast these classical methods with deep learning methods to highlight potential future research directions.

\subsection{Feature Extraction}
We extracted a total of 612 features from all data modalities (see Appendix~\ref{app:baselinefeatures}) across the duration of each video clip in session A and each activity in session B .
Following prior work, we extracted 77 features for ECG and PPG (green channel nose-pad PPG)~\cite{AMIGOS}, 31 features for EDA~\cite{AMIGOS, ASCERTAIN}, and 14 features for respiratory rate~\cite{DEAP}.

Pupil size was inferred over time from the eye-tracking video footage using an open-source eye-tracking algorithm~\cite{JEOresearch_EyeTracker}.
We also computed the mean pixel intensity of each eye for each video frame as a basic visual descriptor.
Additionally, we trained a Fisherface model (PCA followed by LDA)~\cite{belhumeur1997eigenfaces} on each training split for each target variable (affective state, emotion, and personality), and used it to project each video frame into a one-dimensional space.
The resulting per-frame projections (Fisherface features) were included in our analysis.
We used the open-source eye gaze extraction for the Project Aria glasses from Meta~\cite{mansour} to obtain the eye gaze (yaw and pitch). 
As there is no publicly available model yet for blinking detections from Project Aria glasses, we implemented a signal-processing-based approach using the variance map of the eye tracking videos to detect blinks~\cite{Morris}.
To detect micro-expressions from the eye-tracking videos, we extracted features using LBP-TOP~\cite{lbp-top} with a window size of 10~frames (i.e., 111\,ms) similar to previous work on facial videos~\cite{casme2}.
For the acceleration signal from the Aria glasses, we calculated the magnitude across all three axis.
All computations were run on AMD EPYC CPUs. 

For each of the resulting time-series signals---pupil size, eye gaze, video pixel intensity, Fisherface features, and acceleration magnitude---we computed 15 statistical descriptors: mean, minimum, maximum, standard deviation, median, 5th percentile, 95th percentile, range, interquartile range (IQR), sum, energy, skewness, kurtosis, root mean square (RMS), and line integral.
For the micro-expressions, we averaged each of the LBP-TOP features.
The pupil detection and eye-tracking video preprocessing took 2\, hours and about 50\,GB of RAM per participant, with the micro-expressions taking 10 minutes. 
The rest of the feature extraction took under 1\,minute per participant.

\subsection{Continuous Affect Recognition}

For continuous affect recognition, we focused on predicting a participant’s self-reported \emph{arousal}, \emph{valence} and \emph{dominance} levels.
To enable classification, we binarized these continuous ratings into \emph{low} and \emph{high} categories using the median value across the training set, following prior work~\cite{AMIGOS, ASCERTAIN}.

We trained a separate Support Vector Machine (SVM) with a Radial Basis Function (RBF) kernel (default settings~\cite{scikit-learn}) for each target (arousal and valence). 
Features were standardized to the $[0, 1]$ range on a per-participant basis.
No feature selection was applied for this task, as the SVM model was shown to perform robustly with the full set of features.
Classification was performed using a leave-one-subject-out (LOSO) cross-validation strategy to ensure generalization across participants.
We report the mean F1 score across all participants, averaged over the two binary classification tasks (see~\autoref{tab:results} and Appendix \autoref{tab:continuous_affect_prediction_A_B_individual}).

\addtolength{\tabcolsep}{-0.3em}
\begin{table}
    \centering
    \caption{\textbf{Predictions for continuous affect ratings, discrete emotions, and personality traits.}}
    \adjustbox{max width=\textwidth}{
    {\small
        \begin{tabular}{p{1.8cm}p{1.7cm}cccc>{\hspace{1em}}ccccccccc>{\hspace{1em}}c>{\hspace{1em}}c}
        
        \toprule[1.5pt]
         &  & \multicolumn{4}{c}{\textbf{Wearable devices}} & \multicolumn{9}{c}{\textbf{Egocentric glasses}} & \textbf{All} & \textbf{Baseline} \\
         &  &  &  &  &  & \multicolumn{6}{c}{$\overbrace{\hspace{3.9cm}}^{\small{\textnormal{ET video}}}$} &  &  &  &  &  \\
        \textbf{Benchmark}  & \textbf{Domain} & ECG & EDA & RSP & $\bm{\bowtie}$ & Pup. &  Int. & F.f. & Gaze & \hspace{-0.1em}B\hspace{-0.05em}l\hspace{-0.05em}i\hspace{-0.05em}n\hspace{-0.05em}k\hspace{-0.05em} & $\mu$-E. & PPG & IMU & $\bm{\bowtie}$ & $\bm{\bowtie}$ & Random \\
        \midrule
        \midrule
        
        \multirow[c]{3}{*}{\shortstack{\textbf{Continuous} \\ \textbf{Affect}}} & Arousal & 0.76 & 0.76 & 0.75 & 0.76 & 0.77 & 0.76 & 0.76 & \textbf{0.78} & 0.76 & 0.76 & 0.76 & 0.75 & \textbf{0.78} & \textbf{0.78} & 0.64 \\
         & Valence & 0.67 & 0.64 & 0.69 & 0.69 & 0.73 & 0.72 & 0.69 & 0.63 & 0.66 & 0.68 & 0.66 & 0.75  & 0.76 & \textbf{0.77} & 0.55 \\
         & Dominance & 0.63 & 0.66 & 0.66 & 0.66 & 0.67 & 0.66 & 0.65 & 0.65 & \textbf{0.69} & 0.66 & 0.67 & 0.67  & 0.68 & 0.68 & 0.57 \\
         \grayrow & \textbf{Mean} & 0.69 & 0.69 & 0.70 & 0.70 & 0.72 & 0.71 & 0.70 & 0.69 & 0.70 & 0.69 & 0.70 & 0.72 & 0.74 & \textbf{0.75} & 0.59 \\
         
         &  &  &  &  &  &  &  &  &  &  &  & & & & &  \\
        
        \multirow[c]{9}{*}{\shortstack{\textbf{Discrete} \\ \textbf{Emotions}}} & Amused & 0.37 & 0.44 & 0.45 & 0.45 & 0.39 & 0.50 & 0.43 & 0.36 & 0.23 & 0.32 & 0.31 & \textbf{0.58} & 0.50 & 0.52 & 0.21 \\
         & Content & 0.28 & 0.20 & 0.29 & 0.29 & 0.37  & 0.49 & 0.31 & 0.31 & 0.23 & 0.20 & 0.20 & \textbf{0.54} & 0.52 & 0.50 & 0.16 \\
         & Excited & 0.00 & 0.05 & 0.00 & 0.00 & 0.10 & 0.00 & 0.00 & 0.00 & 0.00 & 0.00 & 0.00 &  \textbf{0.12} & \textbf{0.12} & 0.08 & 0.05 \\
         & Awe & 0.00 & 0.00 & 0.00 & 0.00 & 0.24 & 0.00 & 0.00 & 0.06 & 0.05 & 0.00 & 0.00 & 0.23 & \textbf{0.31} & 0.28 & 0.04 \\
         & Neutral & 0.18 & 0.29 & 0.22 & 0.15 & 0.36 & 0.34 & 0.34 & 0.36 & 0.16 & 0.17 & 0.17 & 0.37  & \textbf{0.41} & 0.40 & 0.17 \\
         & Fear & 0.06 & 0.14 & 0.17 & 0.28 & 0.48 & 0.40 & 0.08 & 0.24 & 0.04 & 0.20 & 0.10 & 0.42  & 0.55 & \textbf{0.59} & 0.08 \\
         & Sad & 0.15 & 0.42 & 0.17 & 0.46 & 0.45 &  0.52 & 0.32 & 0.37 & 0.11 & 0.12 & 0.10 & \textbf{0.60} & 0.57 & 0.57 & 0.10 \\
         & Disgust & 0.08 & 0.40 & 0.27 & 0.39 & 0.40 & 0.61 & 0.34 & 0.40 & 0.08 & 0.20 & 0.18 & 0.60 & \textbf{0.65} & 0.61 & 0.12 \\
         & Anger & 0.03 & 0.05 & 0.11 & 0.11 & 0.26 & 0.17 & 0.17 & 0.12 & 0.09 & 0.03 & 0.00 & 0.48 & \textbf{0.53} & 0.50 & 0.08 \\
        \grayrow & \textbf{Mean} & 0.13 & 0.22 & 0.19 & 0.24 & 0.34 & 0.34 & 0.22 & 0.25 & 0.11 & 0.14 & 0.12 & 0.44  & \textbf{0.46} & \textbf{0.46} & 0.11 \\

         &  &  &  &  &  &  &  &  &  &  &  &  & & & &  \\
         
        \multirow[c]{5}{*}{\shortstack{\textbf{Personality} \\ \textbf{Traits}}} & Ex & 0.28 & 0.52 & 0.32 & 0.22 & 0.40 & \textbf{0.60} & 0.50 & 0.58 & 0.43 & 0.48 & 0.30 & 0.55 & 0.55 & 0.48 & 0.55 \\
         & Ag & 0.38 & 0.42 & 0.48 & 0.55 & 0.45 & 0.40 & \textbf{0.60} & \textbf{0.60} & 0.35 & 0.43 & 0.40 & 0.57 & 0.30 & 0.55 & 0.52 \\
         & Co & 0.55 & 0.55 & 0.30 & 0.50 & 0.55 & \textbf{0.65} & 0.45 & 0.48 & 0.58 & 0.55 & 0.57 & 0.40 & 0.55 & \textbf{0.65} & 0.55 \\
         & NE & 0.52 & 0.50 & 0.65 & 0.68 & 0.68 & 0.60 & 0.55 & 0.60 & 0.50 & 0.58 & 0.42 & 0.30 & 0.68 & \textbf{0.70} & 0.52 \\
         & OM & 0.32 & 0.55 & 0.50 & 0.57 & 0.48 & 0.30 & 0.62 & 0.53 & 0.38 & 0.33 & 0.60 & 0.57  & \textbf{0.70} & 0.58 & 0.52 \\
         \grayrow & \textbf{Mean} & 0.41 & 0.51 & 0.45 & 0.50 & 0.51 & 0.51 & 0.54 & 0.56 & 0.45 & 0.47 & 0.46 & 0.48 & 0.57 & \textbf{0.59} & 0.53 \\

        \bottomrule[1.5pt]

        \multicolumn{17}{l}{$\bm{\bowtie}$ = fusion of modalities, Pup. = Pupil size, Int. = Pixel Intensity, F.f. = Fisherface features, $\mu$-E. = micro-expressions. } \\
        \multicolumn{17}{l}{The error bars are not displayed in this table for clarity purposes. They are available in Appendix~\ref{sec:baselines_A_B}. } \\

        \end{tabular}
        }
    }
    \label{tab:results}
    \vspace{-1em}
\end{table}
\addtolength{\tabcolsep}{0.3em}

\subsection{Discrete Emotion Recognition}

In the discrete emotion recognition task, we aimed to classify one of nine basic emotions as reported by participants: \emph{amusement}, \emph{content}, \emph{excitement}, \emph{awe}, \emph{neutral}, \emph{fear}, \emph{sadness}, \emph{disgust}, and \emph{anger}.
For each participant and task, the ground-truth label corresponds to the strongest self-reported emotion.
We used a Random Forest classifier with standardized features.
To reduce dimensionality and focus on the most relevant inputs, we applied SelectKBest~\cite{scikit-learn} feature selection using mutual information, retaining the top 10 features from the training set.
As with the affect recognition task, we employed leave-one-subject-out cross-validation to assess generalization.
We trained a single multi-class classifier and evaluated performance using the mean F1 score across participants (see~\autoref{tab:results} and \autoref{tab:detailed_emotion_prediction_A_B_individual}).
The 9-class classification task had a random baseline F1 score of 0.11.

\subsection{Personality Prediction}

The personality prediction task involves estimating each participant’s Big Five personality traits: \emph{open-mindedness}, \emph{conscientiousness}, \emph{extraversion}, \emph{agreeableness}, and \emph{negative emotionality}.
For each trait, we binarized the self-reported score into \emph{low} and \emph{high} categories using the median across the training set.
We trained a separate Random Forest classifier for each trait.
Unlike the other tasks, we did not apply feature standardization, as the absolute magnitude of certain features was found to be more informative for personality prediction.
For each classifier, we applied SelectKBest feature selection using mutual information and retained the top 10 features from the training set. Feature vectors were constructed by averaging the features for all samples belonging to a participant.
We evaluated the model using leave-one-subject-out cross-validation and report the mean F1 score averaged across all five traits (see~\autoref{tab:results} and Appendix \autoref{tab:personality_prediction_A_B_individual}).

\subsection{Use of deep learning models}

To ground the above results in contemporary approaches, we implemented two deep learning-based models from previous works for wearable emotion recognition: one classical convolutional neural network (CNN)~\cite{DCNN} and one state-of-the-art transformer-based architecture (WER)~\cite{WER}. As input, we use the filtered continuous signals (without Fisherfaces), similar to previous work~\cite{DCNN, WER}.
We trained all models using a five-fold cross-validation approach.
The training was conducted with a batch size of 128 for 30 epochs, a learning rate of 0.0001, and cross-entropy loss as the loss function. Each model was trained on a NVIDIA H200 with a total runtime of about 8 hours for the CNN and 30 hours for the transformer-based architecture.

For all proposed benchmark tasks, our implemented classical methods perform better than the deep learning-based approaches (see Table~\ref{tab:classical_vs_DL}). Using the classical methods, we obtain maximum F1 scores of 0.75 (continuous affect), 0.46 (emotion prediction) and 0.59 (personality prediction) compared to maximum F1 scores of 0.68, 0.23 and 0.47 using the deep learning-based approaches, respectively.
\addtolength{\tabcolsep}{-0.3em}
\begin{table}
    \centering
    \caption{\textbf{Performance comparison between classical and deep learning approaches.}}
    \adjustbox{max width=\textwidth}{
    {\small
        \begin{tabular}{p{2cm}p{2cm}c>{\hspace{1em}}c>{\hspace{1em}}c}
        
        \toprule[1.5pt]
         \multirow[c]{2}{*}{\shortstack{\textbf{Benchmark}}}  & \multirow[c]{2}{*}{\textbf{Model}}  & \multirow[c]{2}{*}{\shortstack{\textbf{Wearable} \\ \textbf{devices}}} & \multirow[c]{2}{*}{\shortstack{\textbf{Egocentric} \\ \textbf{glasses}}} & \multirow[c]{2}{*}{\shortstack{\textbf{All}}} \\
           & & &  &   \\

        \midrule
        \midrule
        
        \multirow[c]{3}{*}{\shortstack{\textbf{Continuous} \\ \textbf{Affect}}} & Classical & 0.70$\pm$0.14 & 0.74$\pm$0.13 & 0.75$\pm$0.13   \\
         & CNN~\cite{DCNN} & 0.63$\pm$0.05  & 0.68$\pm$0.05  & 0.68$\pm$0.07   \\
         & WER~\cite{WER} & 0.49$\pm$0.21  & 0.65$\pm$0.11  & 0.60$\pm$0.16   \\
         
         &  &  &  &      \\
        
        \multirow[c]{3}{*}{\shortstack{\textbf{Discrete} \\ \textbf{Emotions}}} & Classical & 0.24$\pm$0.08  & 0.46$\pm$0.18  & 0.46$\pm$0.17  \\
         & CNN~\cite{DCNN} & 0.12$\pm$0.01  & 0.23$\pm$0.03  & 0.22$\pm$0.02    \\
         & WER~\cite{WER} & 0.13$\pm$0.02  & 0.22$\pm$0.03  & 0.21$\pm$0.04   \\

         &  &  &  &      \\
         
        \multirow[c]{3}{*}{\shortstack{\textbf{Personality} \\ \textbf{Traits}}} & Classical & 0.50$\pm$0.48  & 0.57$\pm$0.49  & 0.59$\pm$0.49 \\
         & CNN~\cite{DCNN} & 0.43$\pm$0.26  & 0.42$\pm$0.20  & 0.41$\pm$0.25     \\
         & WER~\cite{WER} & 0.38$\pm$0.28  & 0.47$\pm$0.24  & 0.44$\pm$0.28     \\

        \bottomrule[1.5pt]

        \end{tabular}
        }
    }
    \label{tab:classical_vs_DL}
    \vspace{-1em}
\end{table}
\addtolength{\tabcolsep}{0.3em}

\subsection{Discussion}

The results in Table~\ref{tab:results} highlight the value of incorporating data from the Aria headset alongside traditional physiological modalities such as ECG, EDA, and RSP.
For continuous affect recognition, the SVM model achieves a mean F1 score of 0.75 when using all modalities.
Signals captured exclusively from the headset reach a comparable mean F1 score of 0.74, slightly outperforming traditional wearable signals ($F_1=$~0.70), with pupil size features contributing strongly.
In the more challenging discrete emotion recognition task, head-mounted signals alone yield a mean F1 score of 0.46, which is substantially above the random baseline of 0.11.
Notably, acceleration magnitude from the Aria IMU achieves $F_1=$~0.44 on its own, and pupil intensity reaches 0.34, while wearable signals perform significantly lower (e.g., $F_1=$~0.13 for ECG, $F_1=$~0.22 for EDA).
For personality prediction, combining all modalities results in a mean F1 score of 0.59 versus 0.53 for the baseline.
Signals from the egocentric glasses alone yield $F_1=$~0.57, outperforming wearable-only inputs ($F_1=$~0.50), with eye gaze being the best-performing individual modality.

These findings suggest that egocentric signals from head-mounted devices, such as eye-tracking video and head motion, capture rich behavioral information beyond traditional physiological sensors.
While such modalities were once impractical in mobile settings, the growing availability of smartglasses and augmented reality headsets makes their use increasingly practical.
Coupled with more expressive models, such as temporal neural networks or multimodal foundation models, these data sources offer promising directions for real-time user state inference and next-generation human-centered systems.
\section{Limitations}
\label{sec:limitations}

While \projdataset offers a rich multimodal dataset for emotion and personality recognition in induced and naturalistic settings, several limitations must be acknowledged. 
First, the ground truth labels rely on retrospective self-reports after each task, which may be affected by recall bias and do not capture the dynamic nature of emotional responses over time~\cite{ G-REx, levenson1988emotion}.
More fine-grained labeling (e.g., via facial expression analysis from our facial recordings) could further improve the temporal resolution of the annotations.
Second, the dataset lacks longitudinal recordings, which may limit the study of emotional and personality state changes over extended periods.
While our study was designed for identifying distinct emotions rather than mapping them to the arousal-valence scale in order to get finer emotion labels, we acknowledge that it has limited representation in the low-arousal, low-valence quadrant. 
We also recognize that some modalities like IMU may primarily capture task-related motor activity rather than affective states due to inherent coupling between behavior and emotion, which could confound emotion recognition with task classification. However, our results show IMU-based prediction performs better in Session A (identical participant behavior across emotions) than in Session B (different participant behavior across emotions), suggesting the IMU captures more than just overt behavioral differences, and is informative for emotion prediction even when behavior is held constant.

Additionally, despite recording eye-tracking and facial video data, we only extracted pupil diameter, pixel intensity, Fisherface features, gaze fixations, blink rate, and micro-expressions. Designing emotion-specific features (e.g., to recognize narrowed eyes when smiling, or teary eyes when sad) would further enhance the performance of the models. Moreover, while we leveraged end-to-end deep learning networks~\cite{li2020deep} in the CNN~\cite{DCNN} and WER~\cite{WER} models we used, we expect that incorporating pre-training on large-scale wearable physiological datasets, as well as future advances in model design and training, will improve the results. We believe our dataset will motivate future research in these directions.
Finally, our participant pool was primarily composed of young adults.
While this may support training stability, it introduces some demographic bias and potentially limits generalization to more diverse populations.

\section{Ethical Considerations and Data Accessibility}
\label{sec:ethicalconsiderations}

The collection of the \projdataset dataset was approved by the ETH Zürich Ethics Commission (no. \texttt{23 ETHICS-008}).
All participants provided informed consent for the recording of their sessions, the creation of the dataset, and its use for research purposes. 
To protect participants' privacy, all personally identifiable information (e.g., age, sex, skin type) and physiological data were anonymized using a numeric participant ID. 
However, given the inherently identifiable nature of egocentric, eye-tracking, and external video data, this information is treated as sensitive.
While emotion and personality recognition can improve mental health monitoring, adaptive interfaces, and user-centric technologies, our dataset could be misused for behavioral profiling or targeted advertising.   
As such, access to this dataset requires users to be permanent staff members of an academic research institution and sign a Data Transfer and Use Agreement to adhere to the terms and conditions of the usage of this dataset. 
The dataset is hosted on servers from ETH Zurich for long-term availability and will be transferred using \textit{sett} (the secure encryption and transfer tool) to minimize the risk of compromised data. Code to analyze the dataset is released under an open-source license. 
\section{Conclusion}

We introduce \projdataset, the first publicly available dataset combining egocentric vision and physiological signals for emotion and personality recognition across both induced and naturalistic tasks. 
Capturing over 50 hours of synchronized multimodal recordings from 43 participants engaged in 16 emotionally diverse activities, \projdataset sets itself apart by covering a broad spectrum of real-world individual and social scenarios.
It proposes three benchmark tasks: continuous affect regression, discrete emotion classification, and personality inference. 
Our results demonstrate that signals from egocentric devices---particularly eye-tracking features and head motion---outperform traditional physiological baselines in emotion and personality recognition tasks. 
These findings highlight the potential of egocentric vision systems to move beyond modeling observable behavior and towards capturing the underlying affective and dispositional states that shape human interaction. 
We envision \projdataset as a foundation for advancing affect-aware human-computer interaction and real-time user state estimation in the wild, enabling more personalized and emotionally intelligent systems across domains such as healthcare, education, and immersive computing.

{
\small
\bibliographystyle{plain} 
\bibliography{sample}

@article{li2020deep,
  title={Deep facial expression recognition: A survey},
  author={Li, Shan and Deng, Weihong},
  journal={IEEE transactions on affective computing},
  volume={13},
  number={3},
  pages={1195--1215},
  year={2020},
  publisher={IEEE}
}

@article{ochsner2000social,
  title={A social cognitive neuroscience approach to emotion and memory},
  author={Ochsner, Kevin N and Schacter, Daniel L},
  journal={The neuropsychology of emotion},
  pages={163--193},
  year={2000}
}

@article{mozikov2024eai,
  title={Eai: Emotional decision-making of llms in strategic games and ethical dilemmas},
  author={Mozikov, Mikhail and Severin, Nikita and Bodishtianu, Valeria and Glushanina, Maria and Nasonov, Ivan and Orekhov, Daniil and Vladislav, Pekhotin and Makovetskiy, Ivan and Baklashkin, Mikhail and Lavrentyev, Vasily and others},
  journal={Advances in Neural Information Processing Systems},
  volume={37},
  pages={53969--54002},
  year={2024}
}

@online{hartig2025multimodal,
  author    = {Pascal Hartig},
  title     = {Building multimodal AI for Ray-Ban Meta glasses},
  year      = {2025},
  month     = mar,
  day       = {4},
  url       = {https://engineering.fb.com/2025/03/04/virtual-reality/building-multimodal-ai-for-ray-ban-meta-glasses/},
  note      = {Meta Engineering Blog}
}

@online{cheng2023mixedreality,
  author    = {Lili Cheng},
  title     = {3 ways mixed reality empowers frontline workers},
  year      = {2023},
  month     = aug,
  day       = {17},
  url       = {https://www.microsoft.com/en-us/industry/blog/manufacturing-and-mobility/2023/08/17/3-ways-mixed-reality-empowers-frontline-workers/},
  note      = {Microsoft Industry Blog}
}

@misc{visionpro,
  author = {Apple Inc.},
  title = {Apple Vision Pro},
  year = {2024},
  howpublished = {\url{https://www.apple.com/apple-vision-pro/}},
  note = {Accessed: 2025-04-13}
}

@misc{quest3,
  author = {Meta},
  title = {Meta Quest 3: Expand your world with Meta Quest 3},
  year = {2024},
  howpublished = {\url{https://www.oculus.com/quest-3/}},
  note = {Accessed: 2025-03-13}
}

@inproceedings{kellnhofer2019gaze360,
  title={Gaze360: Physically unconstrained gaze estimation in the wild},
  author={Kellnhofer, Petr and Recasens, Adria and Stent, Simon and Matusik, Wojciech and Torralba, Antonio},
  booktitle={Proceedings of the IEEE/CVF international conference on computer vision},
  pages={6912--6921},
  year={2019}
}

@inproceedings{ohkawa2023assemblyhands,
  title={Assemblyhands: Towards egocentric activity understanding via 3d hand pose estimation},
  author={Ohkawa, Takehiko and He, Kun and Sener, Fadime and Hodan, Tomas and Tran, Luan and Keskin, Cem},
  booktitle={Proceedings of the IEEE/CVF conference on computer vision and pattern recognition},
  pages={12999--13008},
  year={2023}
}

@inproceedings{kwon2021h2o,
  title={H2o: Two hands manipulating objects for first person interaction recognition},
  author={Kwon, Taein and Tekin, Bugra and St{\"u}hmer, Jan and Bogo, Federica and Pollefeys, Marc},
  booktitle={Proceedings of the IEEE/CVF international conference on computer vision},
  pages={10138--10148},
  year={2021}
}

@inproceedings{garcia2018first,
  title={First-person hand action benchmark with rgb-d videos and 3d hand pose annotations},
  author={Garcia-Hernando, Guillermo and Yuan, Shanxin and Baek, Seungryul and Kim, Tae-Kyun},
  booktitle={Proceedings of the IEEE conference on computer vision and pattern recognition},
  pages={409--419},
  year={2018}
}

@article{russellcircumplex,
  author    = {Russell, James A.},
  title     = {A circumplex model of affect},
  journal   = {Journal of Personality and Social Psychology},
  year      = {1980},
  volume    = {39},
  number    = {6},
  pages     = {1161--1178}
}

@misc{JEOresearch_EyeTracker,
  author       = {JEOresearch},
  title        = {EyeTracker: A lightweight and robust Python eye tracker},
  year         = {2025},
  publisher    = {GitHub},
  howpublished = {\url{https://github.com/JEOresearch/EyeTracker}},
  note         = {Accessed: 2025-03-14}
}

@article{scikit-learn,
  title={Scikit-learn: Machine Learning in {P}ython},
  author={Pedregosa, F. and Varoquaux, G. and Gramfort, A. and Michel, V.
          and Thirion, B. and Grisel, O. and Blondel, M. and Prettenhofer, P.
          and Weiss, R. and Dubourg, V. and Vanderplas, J. and Passos, A. and
          Cournapeau, D. and Brucher, M. and Perrot, M. and Duchesnay, E.},
  journal={Journal of Machine Learning Research},
  volume={12},
  pages={2825--2830},
  year={2011}
}

@article{mikelswheel,
  author    = {Mikels, Joseph A. and Fredrickson, Barbara L. and Larkin, Grace R. and Lindberg, Claire M. and Maglio, Stephen J. and Reuter-Lorenz, Patricia A.},
  title     = {Emotional category data on images from the International Affective Picture System},
  journal   = {Behavior Research Methods},
  year      = {2005},
  volume    = {37},
  number    = {4},
  pages     = {626--630}
}

@article{DEAP,
  author    = {Koelstra, Sander and Muhl, Christian and Soleymani, Mohammad and Lee, Jong-Seok and Yazdani, Ashkan and Ebrahimi, Touradj and Pun, Thierry and Nijholt, Anton and Patras, Ioannis},
  title     = {DEAP: A Database for Emotion Analysis Using Physiological Signals},
  journal   = {IEEE Transactions on Affective Computing},
  volume    = {3},
  number    = {1},
  pages     = {18--31},
  month     = {Jan.-March},
  year      = {2012}
}

@article{ASCERTAIN,
  author    = {Subramanian, R. and Wache, J. and Abadi, M. K. and Vieriu, R. L. and Winkler, S. and Sebe, N.},
  title     = {ASCERTAIN: Emotion and Personality Recognition Using Commercial Sensors},
  journal   = {IEEE Transactions on Affective Computing},
  volume    = {9},
  number    = {2},
  pages     = {147--160},
  month     = {April-June},
  year      = {2018}
}

@article{MAHNHOB-HCI,
  author    = {Soleymani, Mohammad and Lichtenauer, Johan and Pun, Thierry and Pantic, Maja},
  title     = {A Multimodal Database for Affect Recognition and Implicit Tagging},
  journal   = {IEEE Transactions on Affective Computing},
  volume    = {3},
  number    = {1},
  pages     = {42--55},
  month     = {Jan.-March},
  year      = {2012}
}

@article{AMIGOS,
  author    = {Miranda-Correa, J. A. and Abadi, M. K. and Sebe, N. and Patras, I.},
  title     = {AMIGOS: A Dataset for Affect, Personality and Mood Research on Individuals and Groups},
  journal   = {IEEE Transactions on Affective Computing},
  volume    = {12},
  number    = {2},
  pages     = {479--493},
  month     = {April-June},
  year      = {2021}}

@article{K-EMOCON,
  author    = {Park, Chanjoo Y. and Cha, Nayoung and Kang, Sunwoo and Hwang, Hyeoncheol and Cho, Yanghwa and Lee, Joonwon and Lee, Jaewoo and Lee, Jong-Seok},
  title     = {K-EmoCon, a multimodal sensor dataset for continuous emotion recognition in naturalistic conversations},
  journal   = {Scientific Data},
  volume    = {7},
  pages     = {293},
  year      = {2020}}

@article{bfi2,
  author    = {Soto, Christopher J. and John, Oliver P.},
  title     = {The next Big Five Inventory (BFI-2): Developing and assessing a hierarchical model with 15 facets to enhance bandwidth, fidelity, and predictive power},
  journal   = {Journal of Personality and Social Psychology},
  volume    = {113},
  number    = {1},
  pages     = {117-143},
  year      = {2017},
}

@book{costa_bigfive,
  author    = {Costa, P. T. J. and McCrae, R. R.},
  title     = {NEO-PI-R Professional Manual: Revised NEO Personality and NEO Five-Factor Inventory (NEO-FFI)},
  year      = {1992},
  publisher = {Psychological Assessment Resources},
  address   = {Odessa, Florida}
}

@article{Gebhardt,
author = {Gebhardt, Christoph and Brombach, Andreas and Luong, Tiffany and Hilliges, Otmar and Holz, Christian},
title = {Detecting Users' Emotional States during Passive Social Media Use},
year = {2024},
issue_date = {June 2024},
publisher = {Association for Computing Machinery},
address = {New York, NY, USA},
volume = {8},
number = {2},
journal = {Proc. ACM Interact. Mob. Wearable Ubiquitous Technol.},
month = may,
articleno = {77},
numpages = {30}
}

@article{luong2022characterizing,
  title={Characterizing physiological responses to fear, frustration, and insight in virtual reality},
  author={Luong, Tiffany and Holz, Christian},
  journal={IEEE Transactions on Visualization and Computer Graphics},
  volume={28},
  number={11},
  pages={3917--3927},
  year={2022},
  publisher={IEEE}
}

@article{ProjectAria,
  author    = {Jakob Engel and Kiran Somasundaram and Michael Goesele and Albert Sun and Alexander Gamino and Andrew Turner and Arjang Talattof and Arnie Yuan and Bilal Souti and Brighid Meredith and others},
  title     = {Project Aria: A new tool for egocentric multi-modal AI research},
  journal   = {arXiv preprint},
  volume    = {arXiv:2308.13561},
  year      = {2023},
}

@article{picard1997affective,
  author    = {Picard, Rosalind W. and Healey, Jennifer A.},
  title     = {Affective wearables},
  journal   = {Personal Technologies},
  volume    = {1},
  number    = {4},
  pages     = {231--240},
  year      = {1997},
  publisher = {Springer},
}

@article{smets2019into,
  author    = {Smets, E. and De Raedt, W. and Van Hoof, C.},
  title     = {Into the Wild: The Challenges of Physiological Stress Detection in Laboratory and Ambulatory Settings},
  journal   = {IEEE Journal of Biomedical and Health Informatics},
  volume    = {23},
  number    = {2},
  pages     = {463--473},
  year      = {2019},
  month     = {March},
  publisher = {IEEE},
}

@incollection{rottenberg2007emotion,
  author    = {Rottenberg, Jonathan and Ray, Richard D. and Gross, James J.},
  title     = {Emotion Elicitation Using Films},
  booktitle = {Handbook of Emotion Elicitation and Assessment},
  series    = {Series in Affective Science},
  pages     = {12 -- 28},
  publisher = {Oxford University Press},
  year      = {2007}
}

@incollection{levenson1988emotion,
  author    = {Levenson, Robert W.},
  title     = {Emotion and the Autonomic Nervous System: A Prospectus for Research on Autonomic Specificity},
  booktitle = {Social Psychophysiology: Perspectives on Theory and Clinical Applications},
  editor    = {Wagner, H.},
  pages     = {17--42},
  publisher = {Wiley},
  address   = {London},
  year      = {1988}
}

@article{skaramagkas2023esee,
  author    = {Skaramagkas, Vasileios and Ktistakis, Emmanouil and Manousos, Dimitrios and Kazantzaki, Eleni and Tachos, Nikolaos S. and Tripoliti, Evanthia and Fotiadis, Dimitrios I. and Tsiknakis, Manolis},
  title     = {eSEE-d: Emotional State Estimation Based on Eye-Tracking Dataset},
  journal   = {Brain Sciences},
  volume    = {13},
  number    = {4},
  pages     = {589},
  year      = {2023},
  publisher = {MDPI},
}

@article{tarnowski2020eye,
  author    = {Tarnowski, Piotr and Kołodziej, Maciej and Majkowski, Andrzej and Rak, Ryszard J.},
  title     = {Eye‐Tracking Analysis for Emotion Recognition},
  journal   = {Computational Intelligence and Neuroscience},
  volume    = {2020},
  number    = {1},
  pages     = {2909267},
  year      = {2020},
  publisher = {Hindawi},
}

@inproceedings{
das2024emopaircompete,
title={EmoPairCompete - Physiological Signals Dataset for Emotion and Frustration Assessment under Team and Competitive Behaviors},
author={Sneha Das and Nicklas Leander Lund and Carlos Ramos Gonz{\'a}lez and Line H Clemmensen},
booktitle={ICLR 2024 Workshop on Learning from Time Series For Health},
year={2024},
url={https://openreview.net/forum?id=BvgAzJX40Z}
}

@inproceedings{personalityET2019,
author = {Berkovsky, Shlomo and Taib, Ronnie and Koprinska, Irena and Wang, Eileen and Zeng, Yucheng and Li, Jingjie and Kleitman, Sabina},
title = {Detecting Personality Traits Using Eye-Tracking Data},
year = {2019},
isbn = {9781450359702},
publisher = {Association for Computing Machinery},
address = {New York, NY, USA},
url = {https://doi.org/10.1145/3290605.3300451},
doi = {10.1145/3290605.3300451},
booktitle = {Proceedings of the 2019 CHI Conference on Human Factors in Computing Systems},
pages = {1–12},
numpages = {12},
location = {Glasgow, Scotland Uk},
series = {CHI '19}
}

@inproceedings{emoti-sam,
author = {Hayashi, Elaine C. S. and Posada, Juli\'{a}n E. Guti\'{e}rrez and Maike, Vanessa R. M. L. and Baranauskas, M. Cec\'{\i}lia C.},
title = {Exploring new formats of the Self-Assessment Manikin in the design with children},
year = {2016},
isbn = {9781450352352},
publisher = {Association for Computing Machinery},
address = {New York, NY, USA},
url = {https://doi.org/10.1145/3033701.3033728},
booktitle = {Proceedings of the 15th Brazilian Symposium on Human Factors in Computing Systems},
articleno = {27},
numpages = {10},
location = {S\~{a}o Paulo, Brazil},
series = {IHC '16}
}

@inproceedings{yang2021circular,
  author    = {Yang, Jingyuan and Li, Jie and Li, Leida and Wang, Xiumei and Gao, Xinbo},
  title     = {A Circular-Structured Representation for Visual Emotion Distribution Learning},
  booktitle = {Proceedings of the IEEE/CVF Conference on Computer Vision and Pattern Recognition (CVPR)},
  pages     = {4237--4246},
  year      = {2021},
  publisher = {IEEE}
}

@techreport{lang1997iaps,
  author    = {Lang, Peter J. and Bradley, Margaret M. and Cuthbert, Bruce N.},
  title     = {International Affective Picture System (IAPS): Technical Manual and Affective Ratings},
  institution = {NIMH Center for the Study of Emotion and Attention},
  year      = {1997},
  number    = {1},
  pages     = {39--58}
}

@article{hoppe2018eye,
  author    = {Hoppe, Sabrina and Loetscher, Tobias and Morey, Stephanie A. and Bulling, Andreas},
  title     = {Eye Movements During Everyday Behavior Predict Personality Traits},
  journal   = {Frontiers in Human Neuroscience},
  volume    = {12},
  pages     = {105},
  year      = {2018},
  publisher = {Frontiers},
  doi       = {10.3389/fnhum.2018.00105}
}

@inproceedings{wampfler2022,
author = {Wampfler, Rafael and Klingler, Severin and Solenthaler, Barbara and Schinazi, Victor R. and Gross, Markus and Holz, Christian},
title = {Affective State Prediction from Smartphone Touch and Sensor Data in the Wild},
year = {2022},
isbn = {9781450391573},
publisher = {Association for Computing Machinery},
address = {New York, NY, USA},
url = {https://doi.org/10.1145/3491102.3501835},
doi = {10.1145/3491102.3501835},
articleno = {403},
numpages = {14},
location = {New Orleans, LA, USA},
series = {CHI '22}
}

@inproceedings{schmidt2019,
author = {Schmidt, Philip and D\"{u}richen, Robert and Reiss, Attila and Van Laerhoven, Kristof and Pl\"{o}tz, Thomas},
title = {Multi-target affect detection in the wild: an exploratory study},
year = {2019},
isbn = {9781450368704},
publisher = {Association for Computing Machinery},
address = {New York, NY, USA},
url = {https://doi.org/10.1145/3341163.3347741},
doi = {10.1145/3341163.3347741},
booktitle = {Proceedings of the 2019 ACM International Symposium on Wearable Computers},
pages = {211–219},
numpages = {9},
location = {London, United Kingdom},
series = {ISWC '19}
}

@article{G-REx,
  title={A real-world dataset of group emotion experiences based on physiological data},
  author={Bota, Patricia and Brito, Jo{\~a}o and Fred, Ana and others},
  journal={Scientific Data},
  volume={11},
  pages={116},
  year={2024},
  publisher={Nature Publishing Group},
  doi={10.1038/s41597-023-02905-6},
  url={https://doi.org/10.1038/s41597-023-02905-6}
}

@article{biraffe2,
  title={BIRAFFE2, a multimodal dataset for emotion-based personalization in rich affective game environments},
  author={Kutt, Krzysztof and Drążyk, Dominik and Żuchowska, Laura and Szelążek, Michał and Bobek, Szymon and Nalepa, Grzegorz J},
  journal={Scientific Data},
  volume={9},
  pages={274},
  year={2022},
  publisher={Nature Publishing Group},
  doi={10.1038/s41597-022-01402-6},
  url={https://doi.org/10.1038/s41597-022-01402-6}
}

@article{DAPPER,
  title={A dataset of daily ambulatory psychological and physiological recording for emotion research},
  author={Shui, Xuan and Zhang, Meng and Li, Zhihao and others},
  journal={Scientific Data},
  volume={8},
  pages={161},
  year={2021},
  publisher={Nature Publishing Group},
  doi={10.1038/s41597-021-00945-4},
  url={https://doi.org/10.1038/s41597-021-00945-4}
}

@article{PPB-Emo,
  title={A multimodal psychological, physiological and behavioural dataset for human emotions in driving tasks},
  author={Li, Weijian and Tan, Runze and Xing, Yiming and others},
  journal={Scientific Data},
  volume={9},
  pages={481},
  year={2022},
  publisher={Nature Publishing Group},
  doi={10.1038/s41597-022-01557-2},
  url={https://doi.org/10.1038/s41597-022-01557-2}
}

@article{larradet2020toward,
  title={Toward emotion recognition from physiological signals in the wild: Approaching the methodological issues in real-life data collection},
  author={Larradet, François and Niewiadomski, Rafał and Barresi, Giovanni and Caldwell, Darwin G. and Mattos, Leonardo S.},
  journal={Frontiers in Psychology},
  volume={11},
  year={2020},
  publisher={Frontiers Media},
  doi={10.3389/fpsyg.2020.01117},
  url={https://www.frontiersin.org/articles/10.3389/fpsyg.2020.01117}
}

@article{fitzpatrick,
  title={The validity and practicality of sun-reactive skin types I through VI},
  author={Fitzpatrick, Thomas B},
  journal={Archives of Dermatology},
  volume={124},
  number={6},
  pages={869--871},
  year={1988},
  month={Jun},
  publisher={American Medical Association},
  doi={10.1001/archderm.124.6.869},
  pmid={3377516}
}

@article{kwon2021glasses,
  author       = {J. Kwon and J. Ha and D.-H. Kim and J. W. Choi and L. Kim},
  title        = {Emotion Recognition Using a Glasses-Type Wearable Device via Multi-Channel Facial Responses},
  journal      = {IEEE Access},
  volume       = {9},
  pages        = {146392--146403},
  year         = {2021},
  doi          = {10.1109/ACCESS.2021.3121543}
}

@article{SPIDERS,
  title={SPIDERS+: A light-weight, wireless, and low-cost glasses-based wearable platform for emotion sensing and bio-signal acquisition},
  author={Nie, Jingping and Liu, Yanchen and Hu, Yigong and Wang, Yuanyuting and Xia, Stephen and Preindl, Matthias and Jiang, Xiaofan},
  journal={Pervasive and Mobile Computing},
  volume={75},
  pages={101424},
  year={2021},
  publisher={Elsevier},
  doi={10.1016/j.pmcj.2021.101424}
}

@book{IADS,
  title={The International affective digitized sounds (IADS): stimuli, instruction manual and affective ratings},
  author={Bradley, Margaret and Lang, Peter J},
  year={1999},
  publisher={NIMH Center for the Study of Emotion and Attention}
}

@inproceedings{ego4d,
  title={Ego4d: Around the world in 3,000 hours of egocentric video},
  author={Grauman, Kristen and Westbury, Andrew and Byrne, Eugene and Chavis, Zachary and Furnari, Antonino and Girdhar, Rohit and Hamburger, Jackson and Jiang, Hao and Liu, Miao and Liu, Xingyu and others},
  booktitle={Proceedings of the IEEE/CVF conference on computer vision and pattern recognition},
  pages={18995--19012},
  year={2022}
}

@inproceedings{Nymeria,
  title={Nymeria: A massive collection of multimodal egocentric daily motion in the wild},
  author={Ma, Lingni and Ye, Yuting and Hong, Fangzhou and Guzov, Vladimir and Jiang, Yifeng and Postyeni, Rowan and Pesqueira, Luis and Gamino, Alexander and Baiyya, Vijay and Kim, Hyo Jin and others},
  booktitle={European Conference on Computer Vision},
  pages={445--465},
  year={2024},
  organization={Springer}
}

@misc{egoexo4d,
      title={Ego-Exo4D: Understanding Skilled Human Activity from First- and Third-Person Perspectives}, 
      author={Kristen Grauman and Andrew Westbury and Lorenzo Torresani and Kris Kitani and Jitendra Malik and Triantafyllos Afouras and Kumar Ashutosh and Vijay Baiyya and Siddhant Bansal and Bikram Boote et al.},
      year={2024},
      eprint={2311.18259},
      archivePrefix={arXiv},
      primaryClass={cs.CV},
      url={https://arxiv.org/abs/2311.18259}, 
}

@ARTICLE{EPICKITCHENS,
           title={Rescaling Egocentric Vision: Collection, Pipeline and Challenges for EPIC-KITCHENS-100},
           author={Damen, Dima and Doughty, Hazel and Farinella, Giovanni Maria and Furnari, Antonino 
           and Ma, Jian and Kazakos, Evangelos and Moltisanti, Davide and Munro, Jonathan 
           and Perrett, Toby and Price, Will and Wray, Michael},
           journal   = {International Journal of Computer Vision (IJCV)},
           year      = {2022},
           volume = {130},
           pages = {33–55},
           Url       = {https://doi.org/10.1007/s11263-021-01531-2}
}

@article{belhumeur1997eigenfaces,
  title={Eigenfaces vs. fisherfaces: Recognition using class specific linear projection},
  author={Belhumeur, Peter N. and Hespanha, Joao P and Kriegman, David J.},
  journal={IEEE Transactions on pattern analysis and machine intelligence},
  volume={19},
  number={7},
  pages={711--720},
  year={1997},
  publisher={IEEE}
}

@article{braun2025egoppg,
  title={{egoPPG: Heart Rate Estimation from Eye-tracking Cameras in Egocentric Systems to Benefit Downstream Vision Tasks}},
  author={Braun, Bj{\"o}rn and Armani, Rayan and Meier, Manuel and Moebus, Max and Holz, Christian},
  journal={arXiv preprint arXiv:2502.20879},
  year={2025}
}

@article{poh2010non,
  title={Non-contact, automated cardiac pulse measurements using video imaging and blind source separation.},
  author={Poh, Ming-Zher and McDuff, Daniel J and Picard, Rosalind W},
  journal={Optics express},
  volume={18},
  number={10},
  pages={10762--10774},
  year={2010},
  publisher={Optical Society of America}
}

@inproceedings{braun2024suboptimal,
  title={How suboptimal is training rppg models with videos and targets from different body sites?},
  author={Braun, Bj{\"o}rn and McDuff, Daniel and Holz, Christian},
  booktitle={Proceedings of the IEEE/CVF Conference on Computer Vision and Pattern Recognition},
  pages={410--418},
  year={2024}
}

@inproceedings{yu2022physformer,
  title={Physformer: Facial video-based physiological measurement with temporal difference transformer},
  author={Yu, Zitong and Shen, Yuming and Shi, Jingang and Zhao, Hengshuang and Torr, Philip HS and Zhao, Guoying},
  booktitle={Proceedings of the IEEE/CVF conference on computer vision and pattern recognition},
  pages={4186--4196},
  year={2022}
}

@article{braun2023video,
  title={Video-based sympathetic arousal assessment via peripheral blood flow estimation},
  author={Braun, Bj{\"o}rn and McDuff, Daniel and Baltrusaitis, Tadas and Holz, Christian},
  journal={Biomedical Optics Express},
  volume={14},
  number={12},
  pages={6607--6628},
  year={2023},
  publisher={Optica Publishing Group}
}

@inproceedings{braun2024sympcam,
  title={Sympcam: Remote optical measurement of sympathetic arousal},
  author={Braun, Bj{\"o}rn and McDuff, Daniel and Baltrusaitis, Tadas and Streli, Paul and Moebus, Max and Holz, Christian},
  booktitle={2024 IEEE EMBS International Conference on Biomedical and Health Informatics (BHI)},
  pages={1--8},
  year={2024},
  organization={IEEE}
}

@article{moebus2024nightbeat,
  title={Nightbeat: Heart rate estimation from a wrist-worn accelerometer during sleep},
  author={Moebus, Max and Hauptmann, Lars and Kopp, Nicolas and Demirel, Berken and Braun, Bj{\"o}rn and Holz, Christian},
  journal={IEEE Journal of Biomedical and Health Informatics},
  year={2024},
  publisher={IEEE}
}

@article{mansour,
  author={Mansour, Yusuf and Savio Fernandes, Ajoy and Somasundaram, Kiran and Hefny, Tarek and Shakeri, Mahsa and Komogortsev, Oleg V. and Sharma, Abhishek and Proulx, Michael J.},
  journal={IEEE Access}, 
  title={Enabling Eye Tracking for Crowd-Sourced Data Collection With Project Aria}, 
  year={2025},
  volume={13},
  number={},
  pages={114736-114745},
  doi={10.1109/ACCESS.2025.3583623}}

@article{morris,
  author    = {Morris, T. and others},
  title     = {Blink detection for real-time eye tracking},
  journal   = {Journal of Network and Computer Applications},
  year      = {2002}
}

@article{lbp-top,
  author={Zhao, Guoying and Pietikainen, Matti},
  journal={IEEE Transactions on Pattern Analysis and Machine Intelligence}, 
  title={Dynamic Texture Recognition Using Local Binary Patterns with an Application to Facial Expressions}, 
  year={2007},
  volume={29},
  number={6},
  pages={915-928},
  doi={10.1109/TPAMI.2007.1110}}

@article{casme2,
  title={CASME II: An improved spontaneous micro-expression database and the baseline evaluation},
  author={Yan, Wen-Jing and Li, Xiaobai and Wang, Su-Jing and Zhao, Guoying and Liu, Yong-Jin and Chen, Yu-Hsin and Fu, Xiaolan},
  journal={PloS one},
  volume={9},
  number={1},
  pages={e86041},
  year={2014},
  publisher={Public Library of Science San Francisco, USA}
}

@article{kovavcevic2023personality,
  title={Personality trait recognition based on smartphone typing characteristics in the wild},
  author={Kova{\v{c}}evi{\'c}, Nikola and Holz, Christian and G{\"u}nther, Tobias and Gross, Markus and Wampfler, Rafael},
  journal={IEEE Transactions on Affective Computing},
  volume={14},
  number={4},
  pages={3207--3217},
  year={2023},
  publisher={IEEE}
}

@inproceedings{kovacevic,
author = {Kovacevic, Nikola and Holz, Christian and Gross, Markus and Wampfler, Rafael},
title = {On Multimodal Emotion Recognition for Human-Chatbot Interaction in the Wild},
year = {2024},
isbn = {9798400704628},
publisher = {Association for Computing Machinery},
address = {New York, NY, USA},
url = {https://doi.org/10.1145/3678957.3685759},
doi = {10.1145/3678957.3685759},
booktitle = {Proceedings of the 26th International Conference on Multimodal Interaction},
pages = {12–21},
numpages = {10},
location = {San Jose, Costa Rica},
series = {ICMI '24}
}

@article{glabella,
author = {Holz, Christian and Wang, Edward J.},
title = {Glabella: Continuously Sensing Blood Pressure Behavior using an Unobtrusive Wearable Device},
year = {2017},
issue_date = {September 2017},
publisher = {Association for Computing Machinery},
address = {New York, NY, USA},
volume = {1},
number = {3},
url = {https://doi.org/10.1145/3132024},
doi = {10.1145/3132024},
journal = {Proc. ACM Interact. Mob. Wearable Ubiquitous Technol.},
month = sep,
articleno = {58},
numpages = {23}
}

@article{DCNN,
  title={Using deep convolutional neural network for emotion detection on a physiological signals dataset (AMIGOS)},
  author={Santamaria-Granados, Luz and Munoz-Organero, Mario and Ramirez-Gonzalez, Gustavo and Abdulhay, Enas and Arunkumar, NJIA},
  journal={IEEE Access},
  volume={7},
  pages={57--67},
  year={2018},
  publisher={IEEE}
}

@article{WER,
  title={Transformer-based self-supervised multimodal representation learning for wearable emotion recognition},
  author={Wu, Yujin and Daoudi, Mohamed and Amad, Ali},
  journal={IEEE Transactions on Affective Computing},
  volume={15},
  number={1},
  pages={157--172},
  year={2023},
  publisher={IEEE}
}
}

\newpage
\section*{NeurIPS Paper Checklist}

\begin{enumerate}

\item {\bf Claims}
    \item[] Question: Do the main claims made in the abstract and introduction accurately reflect the paper's contributions and scope?
    \item[] Answer: \answerYes{} 
    \item[] Justification: Our claims in the abstract and introduction accurately reflect the paper’s
contributions and scope.
    \item[] Guidelines:
    \begin{itemize}
        \item The answer NA means that the abstract and introduction do not include the claims made in the paper.
        \item The abstract and/or introduction should clearly state the claims made, including the contributions made in the paper and important assumptions and limitations. A No or NA answer to this question will not be perceived well by the reviewers. 
        \item The claims made should match theoretical and experimental results, and reflect how much the results can be expected to generalize to other settings. 
        \item It is fine to include aspirational goals as motivation as long as it is clear that these goals are not attained by the paper. 
    \end{itemize}

\item {\bf Limitations}
    \item[] Question: Does the paper discuss the limitations of the work performed by the authors?
    \item[] Answer: \answerYes{} 
    \item[] Justification: see Section~\ref{sec:limitations}.
    \item[] Guidelines:
    \begin{itemize}
        \item The answer NA means that the paper has no limitation while the answer No means that the paper has limitations, but those are not discussed in the paper. 
        \item The authors are encouraged to create a separate "Limitations" section in their paper.
        \item The paper should point out any strong assumptions and how robust the results are to violations of these assumptions (e.g., independence assumptions, noiseless settings, model well-specification, asymptotic approximations only holding locally). The authors should reflect on how these assumptions might be violated in practice and what the implications would be.
        \item The authors should reflect on the scope of the claims made, e.g., if the approach was only tested on a few datasets or with a few runs. In general, empirical results often depend on implicit assumptions, which should be articulated.
        \item The authors should reflect on the factors that influence the performance of the approach. For example, a facial recognition algorithm may perform poorly when image resolution is low or images are taken in low lighting. Or a speech-to-text system might not be used reliably to provide closed captions for online lectures because it fails to handle technical jargon.
        \item The authors should discuss the computational efficiency of the proposed algorithms and how they scale with dataset size.
        \item If applicable, the authors should discuss possible limitations of their approach to address problems of privacy and fairness.
        \item While the authors might fear that complete honesty about limitations might be used by reviewers as grounds for rejection, a worse outcome might be that reviewers discover limitations that aren't acknowledged in the paper. The authors should use their best judgment and recognize that individual actions in favor of transparency play an important role in developing norms that preserve the integrity of the community. Reviewers will be specifically instructed to not penalize honesty concerning limitations.
    \end{itemize}

\item {\bf Theory assumptions and proofs}
    \item[] Question: For each theoretical result, does the paper provide the full set of assumptions and a complete (and correct) proof?
    \item[] Answer: \answerNA{} 
    \item[] Justification: The paper does not include theoretical results.
    \item[] Guidelines:
    \begin{itemize}
        \item The answer NA means that the paper does not include theoretical results. 
        \item All the theorems, formulas, and proofs in the paper should be numbered and cross-referenced.
        \item All assumptions should be clearly stated or referenced in the statement of any theorems.
        \item The proofs can either appear in the main paper or the supplemental material, but if they appear in the supplemental material, the authors are encouraged to provide a short proof sketch to provide intuition. 
        \item Inversely, any informal proof provided in the core of the paper should be complemented by formal proofs provided in appendix or supplemental material.
        \item Theorems and Lemmas that the proof relies upon should be properly referenced. 
    \end{itemize}

    \item {\bf Experimental result reproducibility}
    \item[] Question: Does the paper fully disclose all the information needed to reproduce the main experimental results of the paper to the extent that it affects the main claims and/or conclusions of the paper (regardless of whether the code and data are provided or not)?
    \item[] Answer: \answerYes{} 
    \item[] Justification: The study protocol is described in section~\ref{sec:dataset} and in further detail in Appendix~\ref{app:study_protocol}. Additionally, we provide the code to run the baselines.
    \item[] Guidelines:
    \begin{itemize}
        \item The answer NA means that the paper does not include experiments.
        \item If the paper includes experiments, a No answer to this question will not be perceived well by the reviewers: Making the paper reproducible is important, regardless of whether the code and data are provided or not.
        \item If the contribution is a dataset and/or model, the authors should describe the steps taken to make their results reproducible or verifiable. 
        \item Depending on the contribution, reproducibility can be accomplished in various ways. For example, if the contribution is a novel architecture, describing the architecture fully might suffice, or if the contribution is a specific model and empirical evaluation, it may be necessary to either make it possible for others to replicate the model with the same dataset, or provide access to the model. In general. releasing code and data is often one good way to accomplish this, but reproducibility can also be provided via detailed instructions for how to replicate the results, access to a hosted model (e.g., in the case of a large language model), releasing of a model checkpoint, or other means that are appropriate to the research performed.
        \item While NeurIPS does not require releasing code, the conference does require all submissions to provide some reasonable avenue for reproducibility, which may depend on the nature of the contribution. For example
        \begin{enumerate}
            \item If the contribution is primarily a new algorithm, the paper should make it clear how to reproduce that algorithm.
            \item If the contribution is primarily a new model architecture, the paper should describe the architecture clearly and fully.
            \item If the contribution is a new model (e.g., a large language model), then there should either be a way to access this model for reproducing the results or a way to reproduce the model (e.g., with an open-source dataset or instructions for how to construct the dataset).
            \item We recognize that reproducibility may be tricky in some cases, in which case authors are welcome to describe the particular way they provide for reproducibility. In the case of closed-source models, it may be that access to the model is limited in some way (e.g., to registered users), but it should be possible for other researchers to have some path to reproducing or verifying the results.
        \end{enumerate}
    \end{itemize}

\item {\bf Open access to data and code}
    \item[] Question: Does the paper provide open access to the data and code, with sufficient instructions to faithfully reproduce the main experimental results, as described in supplemental material?
    \item[] Answer:  \answerYes{} 
    \item[] Justification:  Our paper provides open access to the data and code (both linked to in the abstract), with sufficient instructions to faithfully reproduce the main experimental results described in Sections~\ref{sec:data_analysis} and \ref{sec:baselines}.
    \item[] Guidelines:
    \begin{itemize}
        \item The answer NA means that paper does not include experiments requiring code.
        \item Please see the NeurIPS code and data submission guidelines (\url{https://nips.cc/public/guides/CodeSubmissionPolicy}) for more details.
        \item While we encourage the release of code and data, we understand that this might not be possible, so “No” is an acceptable answer. Papers cannot be rejected simply for not including code, unless this is central to the contribution (e.g., for a new open-source benchmark).
        \item The instructions should contain the exact command and environment needed to run to reproduce the results. See the NeurIPS code and data submission guidelines (\url{https://nips.cc/public/guides/CodeSubmissionPolicy}) for more details.
        \item The authors should provide instructions on data access and preparation, including how to access the raw data, preprocessed data, intermediate data, and generated data, etc.
        \item The authors should provide scripts to reproduce all experimental results for the new proposed method and baselines. If only a subset of experiments are reproducible, they should state which ones are omitted from the script and why.
        \item At submission time, to preserve anonymity, the authors should release anonymized versions (if applicable).
        \item Providing as much information as possible in supplemental material (appended to the paper) is recommended, but including URLs to data and code is permitted.
    \end{itemize}

\item {\bf Experimental setting/details}
    \item[] Question: Does the paper specify all the training and test details (e.g., data splits, hyperparameters, how they were chosen, type of optimizer, etc.) necessary to understand the results?
    \item[] Answer: \answerYes{} 
    \item[] Justification: We specify all details of our employed approach in Section~\ref{sec:baselines}. We use a leave-one-subject-out cross-validation approach, ensuring no data leakage between participants, and use the default settings for all deployed SciPy classifiers. Furthermore, we provide the entire preprocessing, feature calculation, and training and testing pipeline in our code, which can be accessed with the link in the abstract.
    \item[] Guidelines:
    \begin{itemize}
        \item The answer NA means that the paper does not include experiments.
        \item The experimental setting should be presented in the core of the paper to a level of detail that is necessary to appreciate the results and make sense of them.
        \item The full details can be provided either with the code, in appendix, or as supplemental material.
    \end{itemize}

\item {\bf Experiment statistical significance}
    \item[] Question: Does the paper report error bars suitably and correctly defined or other appropriate information about the statistical significance of the experiments?
    \item[] Answer: \answerYes{} 
    \item[] Justification: The paper reports error bars and other statistical significance for the experiments we conducted. We have included such information in the Supplementary Material (Appendix~\ref{sec:baselines_A_B}) to improve clarity in the paper. Additionally, we provide an extensive analysis of the correlations between self-reports and sensor modalities in Appendix~\ref{app:additional_dataset_descriptives}.
    \item[] Guidelines:
    \begin{itemize}
        \item The answer NA means that the paper does not include experiments.
        \item The authors should answer "Yes" if the results are accompanied by error bars, confidence intervals, or statistical significance tests, at least for the experiments that support the main claims of the paper.
        \item The factors of variability that the error bars are capturing should be clearly stated (for example, train/test split, initialization, random drawing of some parameter, or overall run with given experimental conditions).
        \item The method for calculating the error bars should be explained (closed form formula, call to a library function, bootstrap, etc.)
        \item The assumptions made should be given (e.g., Normally distributed errors).
        \item It should be clear whether the error bar is the standard deviation or the standard error of the mean.
        \item It is OK to report 1-sigma error bars, but one should state it. The authors should preferably report a 2-sigma error bar than state that they have a 96\% CI, if the hypothesis of Normality of errors is not verified.
        \item For asymmetric distributions, the authors should be careful not to show in tables or figures symmetric error bars that would yield results that are out of range (e.g. negative error rates).
        \item If error bars are reported in tables or plots, The authors should explain in the text how they were calculated and reference the corresponding figures or tables in the text.
    \end{itemize}

\item {\bf Experiments compute resources}
    \item[] Question: For each experiment, does the paper provide sufficient information on the computer resources (type of compute workers, memory, time of execution) needed to reproduce the experiments?
    \item[] Answer: \answerYes{}
    \item[] We provide all compute details in Section~\ref{sec:baselines}. All computations were run on AMD EPYC CPUs. One CPU is enough to process each participant individually, with about 50\,GB of RAM necessary. 
    Processing of the pupils and video data took about 2\,hours per participant.
    The rest of the feature extraction and training/testing took under 1 minute per participant and feature.
    \item[] Guidelines:
    \begin{itemize}
        \item The answer NA means that the paper does not include experiments.
        \item The paper should indicate the type of compute workers CPU or GPU, internal cluster, or cloud provider, including relevant memory and storage.
        \item The paper should provide the amount of compute required for each of the individual experimental runs as well as estimate the total compute. 
        \item The paper should disclose whether the full research project required more compute than the experiments reported in the paper (e.g., preliminary or failed experiments that didn't make it into the paper). 
    \end{itemize}
    
\item {\bf Code of ethics}
    \item[] Question: Does the research conducted in the paper conform, in every respect, with the NeurIPS Code of Ethics \url{https://neurips.cc/public/EthicsGuidelines}?
    \item[] Answer: \answerYes{} 
    \item[] Justification: The research conducted in the paper conform, in every respect, with the NeurIPS Code of Ethics. 

    \item[] Guidelines:
    \begin{itemize}
        \item The answer NA means that the authors have not reviewed the NeurIPS Code of Ethics.
        \item If the authors answer No, they should explain the special circumstances that require a deviation from the Code of Ethics.
        \item The authors should make sure to preserve anonymity (e.g., if there is a special consideration due to laws or regulations in their jurisdiction).
    \end{itemize}

\item {\bf Broader impacts}
    \item[] Question: Does the paper discuss both potential positive societal impacts and negative societal impacts of the work performed?
    \item[] Answer: \answerYes{} 
    \item[] Justification: The paper discuss both potential positive societal impacts and negative societal impacts of the work performed in section~\ref{sec:ethicalconsiderations}.
    \item[] Guidelines:
    \begin{itemize}
        \item The answer NA means that there is no societal impact of the work performed.
        \item If the authors answer NA or No, they should explain why their work has no societal impact or why the paper does not address societal impact.
        \item Examples of negative societal impacts include potential malicious or unintended uses (e.g., disinformation, generating fake profiles, surveillance), fairness considerations (e.g., deployment of technologies that could make decisions that unfairly impact specific groups), privacy considerations, and security considerations.
        \item The conference expects that many papers will be foundational research and not tied to particular applications, let alone deployments. However, if there is a direct path to any negative applications, the authors should point it out. For example, it is legitimate to point out that an improvement in the quality of generative models could be used to generate deepfakes for disinformation. On the other hand, it is not needed to point out that a generic algorithm for optimizing neural networks could enable people to train models that generate Deepfakes faster.
        \item The authors should consider possible harms that could arise when the technology is being used as intended and functioning correctly, harms that could arise when the technology is being used as intended but gives incorrect results, and harms following from (intentional or unintentional) misuse of the technology.
        \item If there are negative societal impacts, the authors could also discuss possible mitigation strategies (e.g., gated release of models, providing defenses in addition to attacks, mechanisms for monitoring misuse, mechanisms to monitor how a system learns from feedback over time, improving the efficiency and accessibility of ML).
    \end{itemize}
    
\item {\bf Safeguards}
    \item[] Question: Does the paper describe safeguards that have been put in place for responsible release of data or models that have a high risk for misuse (e.g., pretrained language models, image generators, or scraped datasets)?
    \item[] Answer: \answerYes{} 
    \item[] Justification: The paper describes in section~\ref{sec:ethicalconsiderations} the safeguards that have been put in place for responsible release of data that have a high risk for misuse.
    \item[] Guidelines:
    \begin{itemize}
        \item The answer NA means that the paper poses no such risks.
        \item Released models that have a high risk for misuse or dual-use should be released with necessary safeguards to allow for controlled use of the model, for example by requiring that users adhere to usage guidelines or restrictions to access the model or implementing safety filters. 
        \item Datasets that have been scraped from the Internet could pose safety risks. The authors should describe how they avoided releasing unsafe images.
        \item We recognize that providing effective safeguards is challenging, and many papers do not require this, but we encourage authors to take this into account and make a best faith effort.
    \end{itemize}

\item {\bf Licenses for existing assets}
    \item[] Question: Are the creators or original owners of assets (e.g., code, data, models), used in the paper, properly credited and are the license and terms of use explicitly mentioned and properly respected?
    \item[] Answer: \answerYes{} 
    \item[] Justification: The original owners of code, dataset and model used in our paper are properly cited and credited. The license and terms of use explicitly mentioned in the supplemental material and properly respected.
    \item[] Guidelines:
    \begin{itemize}
        \item The answer NA means that the paper does not use existing assets.
        \item The authors should cite the original paper that produced the code package or dataset.
        \item The authors should state which version of the asset is used and, if possible, include a URL.
        \item The name of the license (e.g., CC-BY 4.0) should be included for each asset.
        \item For scraped data from a particular source (e.g., website), the copyright and terms of service of that source should be provided.
        \item If assets are released, the license, copyright information, and terms of use in the package should be provided. For popular datasets, \url{paperswithcode.com/datasets} has curated licenses for some datasets. Their licensing guide can help determine the license of a dataset.
        \item For existing datasets that are re-packaged, both the original license and the license of the derived asset (if it has changed) should be provided.
        \item If this information is not available online, the authors are encouraged to reach out to the asset's creators.
    \end{itemize}

\item {\bf New assets}
    \item[] Question: Are new assets introduced in the paper well documented and is the documentation provided alongside the assets?
    \item[] Answer: \answerYes{} 
    \item[] Justification: All new assets introduced in our paper including, data, code, and models are well documented with documents included in the paper (links available in abstract).
    \item[] Guidelines:
    \begin{itemize}
        \item The answer NA means that the paper does not release new assets.
        \item Researchers should communicate the details of the dataset/code/model as part of their submissions via structured templates. This includes details about training, license, limitations, etc. 
        \item The paper should discuss whether and how consent was obtained from people whose asset is used.
        \item At submission time, remember to anonymize your assets (if applicable). You can either create an anonymized URL or include an anonymized zip file.
    \end{itemize}

\item {\bf Crowdsourcing and research with human subjects}
    \item[] Question: For crowdsourcing experiments and research with human subjects, does the paper include the full text of instructions given to participants and screenshots, if applicable, as well as details about compensation (if any)? 
    \item[] Answer: \answerYes{} 
    \item[] Justification: The paper includes the full set of instructions given to participants as well as the compensation received by participants (see section~\ref{sec:dataset} and Supplementary Material).
    \item[] Guidelines:
    \begin{itemize}
        \item The answer NA means that the paper does not involve crowdsourcing nor research with human subjects.
        \item Including this information in the supplemental material is fine, but if the main contribution of the paper involves human subjects, then as much detail as possible should be included in the main paper. 
        \item According to the NeurIPS Code of Ethics, workers involved in data collection, curation, or other labor should be paid at least the minimum wage in the country of the data collector. 
    \end{itemize}

\item {\bf Institutional review board (IRB) approvals or equivalent for research with human subjects}
    \item[] Question: Does the paper describe potential risks incurred by study participants, whether such risks were disclosed to the subjects, and whether Institutional Review Board (IRB) approvals (or an equivalent approval/review based on the requirements of your country or institution) were obtained?
    \item[] Answer: \answerYes{} 
    \item[] Justification: The study protocol was approved by the ETH Zürich Ethics committee (\texttt{no. 23 ETHICS-008)}, as described in section~\ref{sec:ethicalconsiderations}.
    \item[] Guidelines:
    \begin{itemize}
        \item The answer NA means that the paper does not involve crowdsourcing nor research with human subjects.
        \item Depending on the country in which research is conducted, IRB approval (or equivalent) may be required for any human subjects research. If you obtained IRB approval, you should clearly state this in the paper. 
        \item We recognize that the procedures for this may vary significantly between institutions and locations, and we expect authors to adhere to the NeurIPS Code of Ethics and the guidelines for their institution. 
        \item For initial submissions, do not include any information that would break anonymity (if applicable), such as the institution conducting the review.
    \end{itemize}

\item {\bf Declaration of LLM usage}
    \item[] Question: Does the paper describe the usage of LLMs if it is an important, original, or non-standard component of the core methods in this research? Note that if the LLM is used only for writing, editing, or formatting purposes and does not impact the core methodology, scientific rigorousness, or originality of the research, declaration is not required.
    \item[] Answer: \answerNA{} 
    \item[] Justification: The core method development in this research does not involve LLMs as any important, original, or non-standard components.
    \item[] Guidelines:
    \begin{itemize}
        \item The answer NA means that the core method development in this research does not involve LLMs as any important, original, or non-standard components.
        \item Please refer to our LLM policy (\url{https://neurips.cc/Conferences/2025/LLM}) for what should or should not be described.
    \end{itemize}

\end{enumerate}

\newpage

\appendix
\section*{Appendix}

\section{Study Protocol}
\label{app:study_protocol}

\begin{figure}[H]
  \centering
  \includegraphics[width=\textwidth]{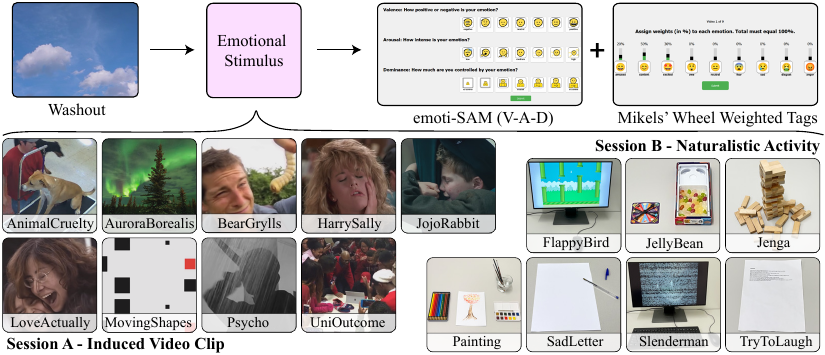}
  \caption{\textbf{Overview of the experimental protocol}. The experiment consisted of two sessions. In session A, participants watched 9 video clips, with a 40 s washout between clips and a 5 s video of a cross preceding each clip. 
  In session B, participants performed 7 real-world tasks. 
  Each task was spaced by a 1-min washout clip. 
  Two questionnaires, corresponding to the emoti-SAM~\cite{emoti-sam} and a weighted Mikels' Wheel \cite{mikelswheel} were answered after each emotional stimulus.} 
  \label{fig:study_protocol}
\end{figure}

Before starting the experiment, an explanation of the study protocol, illustrated in Figure~\ref{fig:study_protocol}, was given to each participant.
The experiment consisted in participants watching 9 videos and performing 7 tasks, as listed in Table~\ref{tab:naturalistic_activities}. Details of the videos are provided in Table~\ref{tab:description_video_clips}.
We informed participants that each target emotion would be experienced only once during session A. 
This ensured that, after viewing a disturbing video, such as one eliciting disgust or fear, they would not anticipate encountering a similar emotional stimulus in the remaining videos.
Between each stimulus, a washout video of clouds was shown to mitigate any emotional carry-over effect. 
Washouts lasted 40 seconds in session A and 1 minute in session B.  

\begin{figure}[H]
  \centering
  \includegraphics[width=\textwidth]{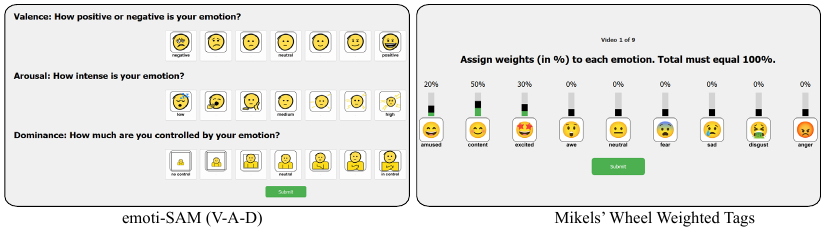}
  \caption{\textbf{Close-up view of the self-report questionnaires}. In the emoti-SAM~\cite{emoti-sam}, participants rated their arousal, valence and dominance using a 7-point scale. In the weighted Mikels' Wheel \cite{mikelswheel}, participants distributed a 100\% weight across emotions in 10\% increments.} 
  \vspace{-2em}
  \label{fig:selfreports}
\end{figure}

Following each emotional stimulus, participants rated their emotions using an emoti-SAM~\cite{emoti-sam} and a weighted Mikels' Wheel~\cite{mikelswheel}, as shown in Figure~\ref{fig:selfreports}. 
To familiarise the participant with each questionnaire, we explained what each term in the emoti-SAM meant i.e., \emph{arousal}, \emph{valence}, \emph{dominance} and provided a definition for each emotion on Mikels' Wheel. 
In addition, we gave two examples of emotions and their associated self-reports. 
For the weighted Mikels' Wheel questionnaire, we indicated to the participant that they could gauge the intensity of their emotion using the neutral emotion. 
For example, if only feeling a single emotion but in low intensity, the participant could distribute the remaining weights in the neutral emotion. (e.g., 20\% amused and 80\% neutral indicates low amusement).

\newpage
\section{Additional Dataset Descriptives}
\label{app:additional_dataset_descriptives}

\subsection{Mean self-ratings per task}
\begin{figure}[H]
  \centering
  \begin{subfigure}{\textwidth}
    \centering
    \includegraphics[width=\linewidth]{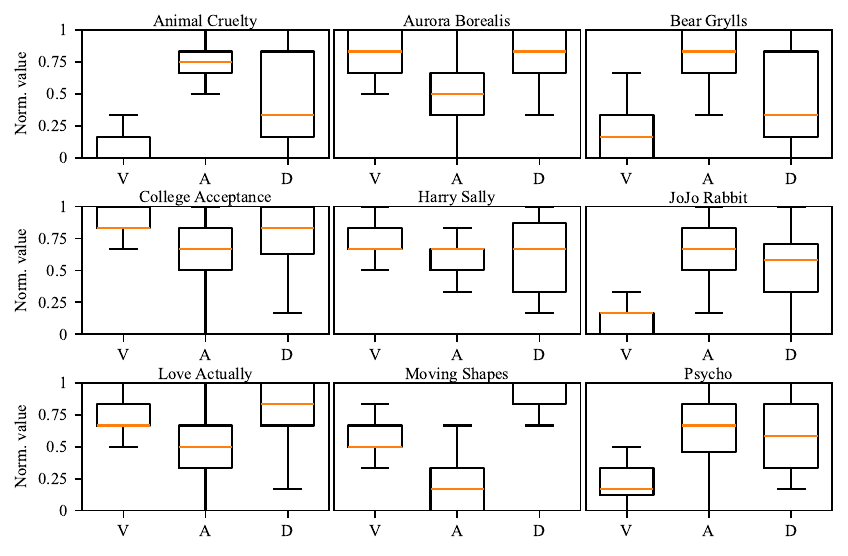}
      \vspace{-2em} 
    \caption{}
    \label{fig:mean_self_ratings_A}
  \end{subfigure}%

  \begin{subfigure}{\textwidth}
    \centering
    \includegraphics[width=\linewidth]{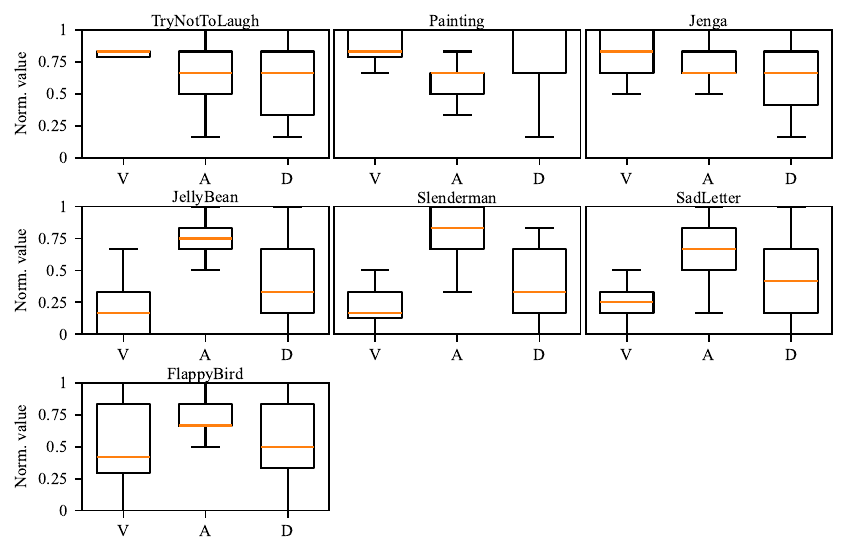}
      \vspace{-2em} 
    \caption{}
    \label{fig:mean_self_ratings_B}
  \end{subfigure}
  \caption{Box plots of affect self-ratings \textbf{a)} per video clip \textbf{b)} per naturalistic task.}
\end{figure}

The normalized continuous affect self-ratings for all video clips, averaged across participants, is displayed in Figure~\ref{fig:mean_self_ratings_A}.
Similarly, the mean continuous affect self-ratings for the naturalistic activities of session B are displayed in Figure~\ref{fig:mean_self_ratings_B}. 

\subsection{Self-correlations of continuous self-ratings}

\begin{figure}[H]
    \centering
     \includegraphics[width=\linewidth]{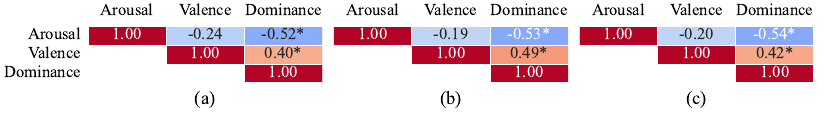}
    \caption{Pearson correlations between continuous self-ratings in \textbf{a)} session A \textbf{b)}  session B \textbf{c)} session A+B.}
    \label{fig:self_correlation_affet_domains}
\end{figure}

Figure~\ref{fig:self_correlation_affet_domains} presents the Pearson correlation matrices between continuous affect self-ratings (A-V-D) across sessions A, B and A+B. 
Across all sessions, a strong negative relationship between arousal and dominance was observed, as well as a moderate positive relationship between valence and dominance. 
This indicated participants associated intense emotions with  low dominance and vice-versa, while associating negative emotions to low dominance.

\subsection{Self-correlations of discrete emotions}

\begin{figure}[H]
  \centering
    \includegraphics{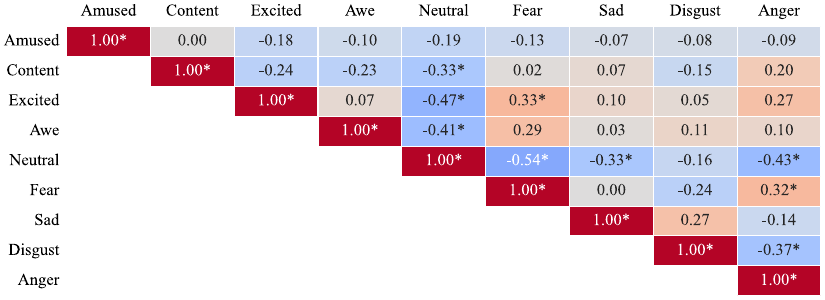}
  \caption{Pearson correlations between discrete emotions (session A).}
    \label{fig:self_correlation_emotions_a}
\end{figure}

\begin{figure}[H]
  \centering
    \includegraphics{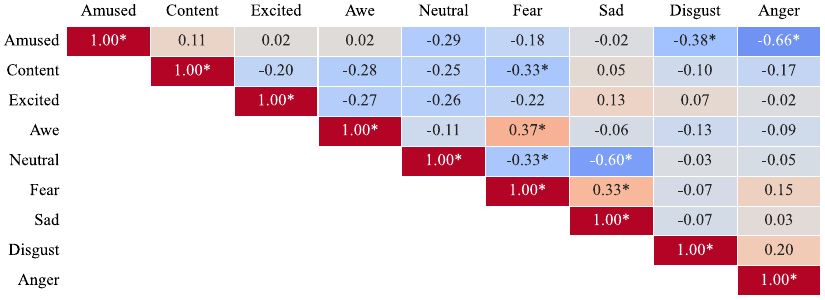}
  \caption{Pearson correlations between discrete emotions (session B).}
    \label{fig:self_correlation_emotions_b}
\end{figure}

\begin{figure}[H]
  \centering
    \includegraphics{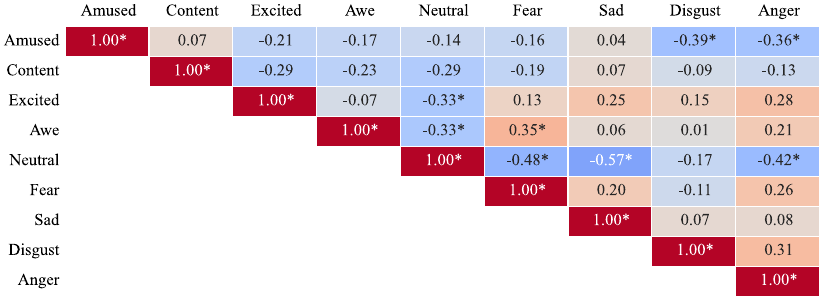}
  \caption{Pearson correlations between discrete emotions (session A+B).}
    \label{fig:self_correlation_emotions_a_and_b}
\end{figure}

Figures~\ref{fig:self_correlation_emotions_a}, ~\ref{fig:self_correlation_emotions_b}, ~\ref{fig:self_correlation_emotions_a_and_b} present the Pearson correlation matrices between discrete emotions across sessions A, B and A+B, respectively.
The video clips of session A resulted in significant negative relationships between the neutral emotions and all other emotions excluding amused and disgust.
Fear had a positive correlation with excitement and anger, while anger had a negative relationship with disgust. 
In session B, amusement had a strong negative correlation with anger and a moderate negative correlation with disgust. 

Fear had a moderate positive relationship with content and neutral, while having a negative relationship with awe and sadness.
Finally, sadness had a strong positive relationship with the neutral emotion.

After combining the self-reports of discrete emotions from session A and B, amusement was negatively correlated with disgust and anger, while fear was positively correlated with awe.
The neutral emotion was negatively correlated with excitement, awe, fear, sadness and anger.

\subsection{Self-correlations of personality traits}

\begin{figure}[H]
  \centering
    \includegraphics[width=0.5\textwidth]{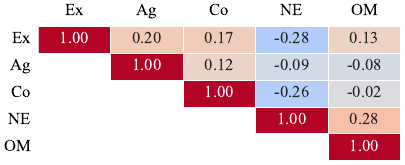}
  \caption{Pearson correlations between personality traits.}
    \label{fig:self_correlation_personality}
\end{figure}

Figure~\ref{fig:self_correlation_personality} presents the Pearson correlations between personality traits. No significant correlation was found between personality traits.

\subsection{Correlations between continuous self-ratings and personality traits.}

\begin{figure}[H]
  \centering
  \begin{subfigure}{0.45\textwidth}
    \centering
    \includegraphics[width=\linewidth]{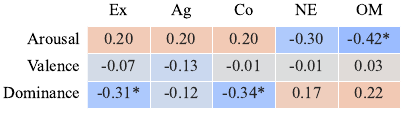}
    \caption{}
    \label{fig:correlation_continuous_personality_A}
  \end{subfigure}%
  \hspace{0.5em}
  \begin{subfigure}{0.45\textwidth}
    \centering
    \includegraphics[width=\linewidth]{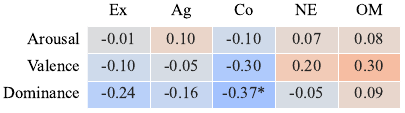}
    \caption{}
    \label{fig:correlation_continuous_personality_B}
  \end{subfigure}
  \caption{Pearson correlations between continuous self-ratings and personality scores in \textbf{a)} session A \textbf{b)} session B.}
\end{figure}

Figures~\ref{fig:correlation_continuous_personality_A} and ~\ref{fig:correlation_continuous_personality_B} display the correlations between the continuous self-ratings and personality traits in session A and session B, respectively.
In session A, significant negative correlations are found between dominance and \emph{Ex} and \emph{Co}, as well as arousal and \emph{OM}.
In session B, a negative relationship between dominance and \emph{Co} was observed.

\subsection{Correlations between continuous self-ratings and discrete emotions.}

\begin{figure}[H]
  \centering
  \begin{subfigure}{\textwidth}
    \centering
    \includegraphics[width=\linewidth]{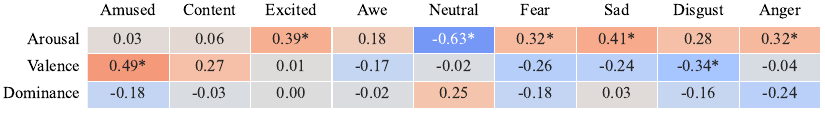}
    \caption{}
    \label{fig:correlation_continuous_emotions_A}
  \end{subfigure}%
  \vspace{1em} 
  \begin{subfigure}{\textwidth}
    \centering
    \includegraphics[width=\linewidth]{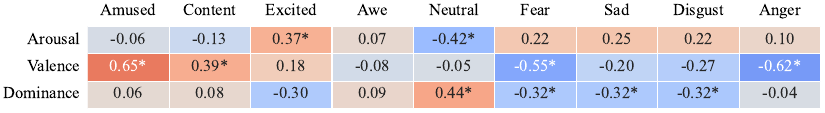}
    \caption{}
    \label{fig:correlation_continuous_emotions_B}
  \end{subfigure}
  \caption{Pearson correlations between continuous self-ratings and discrete emotions in \textbf{a)} session A \textbf{b)} session B.}
\end{figure}

Figures~\ref{fig:correlation_continuous_emotions_A} and \ref{fig:correlation_continuous_emotions_B} present the Pearson correlations between continuous self-ratings and discrete emotions in session A and session B, respectively.
In session A, arousal was positively correlated with excitement, fear, sadness and anger, while being negatively correlated with the neutral emotion. Valence was positively correlated with amusement and negatively correlated with disgust.
In session B, arousal was positively correlated with excitement and negatively correlated with the neutral emotion. 
Valence was strongly positively correlated to amusement, while being strongly negatively correlated with fear and anger. 
Dominance was negatively correlated with fear, sadness and disgust, while being positively correlated with the neutral emotion.

\subsection{Pearson correlations between personality scores and discrete emotions.}

\begin{figure}[H]
  \centering
  \begin{subfigure}{\textwidth}
    \centering
    \includegraphics[width=\linewidth]{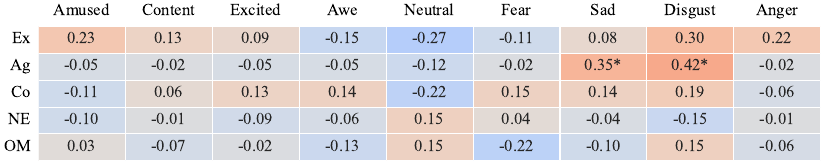}
    \caption{}
    \label{fig:correlation_personality_emotions_A}
  \end{subfigure}%
  \vspace{1em} 
  \begin{subfigure}{\textwidth}
    \centering
    \includegraphics[width=\linewidth]{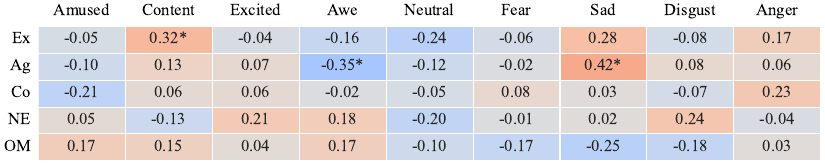}
    \caption{}
    \label{fig:correlation_personality_emotions_B}
  \end{subfigure}
  \caption{Pearson correlation between personality scores and discrete emotions in \textbf{a)} session A \textbf{b)} session B.}
\end{figure}

Figures~\ref{fig:correlation_personality_emotions_A} and \ref{fig:correlation_personality_emotions_B} present the Pearson correlations between personality traits and discrete emotions in session A and session B, respectively.
In session A, significant positive correlations were observed between \emph{Ag} and sadness and disgust.
In session B, \emph{Ex} was positively correlated with content. \emph{Ag} was positively correlated with sadness and negatively correlated with awe.

\section{Additional Descriptions}
\label{sec:baselines_A_B}

\subsection{Detailed Description of Baseline Features}
\label{app:baselinefeatures}
 A detailed overview of the features extracted from each modality is presented in Table~\ref{tab:modality_features}. 

\begin{table}[H]
\centering
\caption{\textbf{Overview of features extracted from the recorded physiological time-series signals.} Recorded physiological signals include respiration rate, ECG, EDA, PPG recorded using a nosepad sensor and acceleration magnitude from the Aria-integrated IMU sensor. We further compute statistical descriptors of the time-series signals inferred from the video data captured by the eye-tracking cameras, including pupil size, video pixel intensity, and Fisherface feature coefficients.}
\resizebox{\textwidth}{!}{%
\begin{tabular}{lp{10cm}}
\toprule[1.5pt]
\textbf{Time-series signal} & \textbf{Features extracted} \\
\midrule
\midrule
Acceleration Magnitude (Aria IMU) & Mean, min, max, standard deviation, median, 5th and 95th percentiles, range, interquartile range, sum, energy, skewness, kurtosis, RMS, and line integral. \\
\hline
Blinking (ET camera) & 
Number of blinks of the left eye and the right eye.
\\
\hline
ECG & 
Root mean square of the mean squared IBIs, mean IBI, 60 spectral power values in the [0–6] Hz band of the ECG signal, low-frequency [0.01–0.08] Hz, medium-frequency [0.08–0.15] Hz, and high-frequency [0.15–0.5] Hz components of HRV spectral power, HR and HRV statistics. \\
\hline
EDA & 
Mean skin resistance and mean of derivative, mean differential for negative values only (mean decrease rate during decay time), proportion of negative derivative samples, number of local minima in the GSR signal, average rising time of the GSR signal, spectral power in the [0–2.4] Hz band, zero crossing rate of skin conductance slow response (SCSR) [0–0.2] Hz, zero crossing rate of skin conductance very slow response (SCVSR) [0–0.08] Hz, mean SCSR and SCVSR peak magnitude. \\
\hline
Eye Gaze (Yaw and Pitch) & 
Mean, min, max, standard deviation, median, 5th and 95th percentiles, range, interquartile range, sum, energy, skewness, kurtosis, RMS, and line integral.
\\
\hline
Fisherface Features (ET camera) & Mean, min, max, standard deviation, median, 5th and 95th percentiles, range, interquartile range, sum, energy, skewness, kurtosis, RMS, and line integral.\\
\hline
Micro-expressions (ET camera) & 
Mean of each LBP-TOP feature.
\\
\hline
PPG (Nosepad Sensor) & Root mean square of the mean squared IBIs, mean IBI, 60 spectral power values in the [0–6] Hz band of the PPG signal, low-frequency [0.01–0.08] Hz, medium-frequency [0.08–0.15] Hz, and high-frequency [0.15–0.5] Hz components of HRV spectral power, HR and HRV statistics. \\
\hline
Pupil Size (ET camera) & Mean, min, max, standard deviation, median, 5th and 95th percentiles, range, interquartile range, sum, energy, skewness, kurtosis, RMS, and line integral. \\
\hline
RSP & 
Band energy ratio (difference between the logarithm of energy between the lower (0.05–0.25 Hz) and the higher (0.25–5 Hz) bands), average respiration signal, mean of derivative (variation of the respiration signal), standard deviation, range or greatest breath, breathing rhythm (spectral centroid), breathing rate, 10 spectral power values in the bands from 0 to 2.4 Hz, average peak-to-peak time, median peak-to-peak time. \\
\hline
Video Pixel Intensity (ET camera) & Mean, min, max, standard deviation, median, 5th and 95th percentiles, range, interquartile range, sum, energy, skewness, kurtosis, RMS, and line integral. \\
\bottomrule[1.5pt]
\end{tabular}
}
\label{tab:modality_features}
\end{table}

\subsection{Continuous Affect Prediction}

Table~\ref{tab:continuous_affect_prediction_A_B_individual} presents the continuous affect domain prediction results for session A and session B.
The egocentric glasses provided better predictions of the continuous affect self-reports than the physiological sensors.
The glasses had a $F_1$ score of 0.72 and 0.73 in session A and B, respectively, while the physiological sensors reported an $F_1$ score of 0.68 in session A and session B. 
Notably, all sensors had a strong performance when predicting arousal in session B. 

Table~\ref{tab:continuous_affect_prediction_STD} presents the standard deviation results of the continuous affect domain predictions across all sessions.
Session A displays greater variability, particularly within the arousal and dominance self-reports. 
In contrast, Session B demonstrates lower and more uniform standard deviations across all modalities.
When aggregating session A with session B, the lowest standard deviations are achieved.

\addtolength{\tabcolsep}{-0.3em}
\begin{table}[H]
    \centering
    \caption{\textbf{Continuous affect domain prediction results.}}
    \adjustbox{max width=\textwidth}{

     {\small
        \begin{tabular}{p{1.8cm}p{1.7cm}cccc>{\hspace{1em}}ccccccccc>{\hspace{1em}}c>{\hspace{1em}}c}
        
        \toprule[1.5pt]
         &  & \multicolumn{4}{c}{\textbf{Wearable devices}} & \multicolumn{9}{c}{\textbf{Egocentric glasses}} & \textbf{All} & \textbf{Baseline} \\
         &  &  &  &  &  & \multicolumn{6}{c}{$\overbrace{\hspace{3.9cm}}^{\small{\textnormal{ET video}}}$} &  &  &  &  &  \\
        Session  & Domain & ECG & EDA & RSP & $\bm{\bowtie}$ & Pup. &  Int. & F.f. & Gaze & \hspace{-0.1em}B\hspace{-0.05em}l\hspace{-0.05em}i\hspace{-0.05em}n\hspace{-0.05em}k\hspace{-0.05em} & $\mu$-E. & PPG & IMU & $\bm{\bowtie}$ & $\bm{\bowtie}$ & Random \\
        \midrule
        \midrule
        \multirow[c]{3}{*}{\textbf{A}} & Arousal & 0.64 & 0.65 & 0.60 & 0.63 & 0.67 & 0.62 & 0.64   & 0.69 & 0.66 & 0.64 & 0.64 & 0.65 & 0.68 & \textbf{0.69} & 0.56 \\
         & Valence & 0.71 & 0.68 & 0.62 & 0.71 & 0.71 & 0.66 & \textbf{0.78}                        & 0.64 & 0.71 & 0.63 & 0.77 & 0.66 & \textbf{0.78} & \textbf{0.78} & 0.56 \\
         & Dominance & 0.70 & 0.70 & 0.71 & 0.71 & 0.68 & \textbf{0.72} & 0.68                      & 0.71 & 0.72 & 0.71 & 0.69 & 0.71 & 0.71 & 0.71 & 0.61 \\
        \grayrow & \textbf{Mean} & 0.68 & 0.68 & 0.65 & 0.68 & 0.69 & 0.67 & 0.70                   & 0.68 & 0.70 & 0.66 & 0.71 & 0.68 & 0.72 & \textbf{0.73} & 0.58 \\
         
         &  &  &  &  &  &  &  &  &  &  &  &  &  \\
        
         \multirow[c]{3}{*}{\textbf{B}} & Arousal & \textbf{0.83} & \textbf{0.83} & \textbf{0.83} & \textbf{0.83} & \textbf{0.83} &\textbf{0.83} & \textbf{0.83} & \textbf{0.83} & \textbf{0.83} & \textbf{0.83} & \textbf{0.83} & \textbf{0.83} & \textbf{0.83} & \textbf{0.83} & 0.74 \\
         & Valence & 0.64 & 0.57 & 0.62 & 0.68 & 0.72 & 0.66 & \textbf{0.78}                & 0.66 & 0.70 & 0.67 & 0.71 & 0.71 & 0.77 & 0.72 & 0.53 \\
         & Dominance & 0.51 & 0.50 & 0.55 & 0.54 & 0.56 & 0.55 & 0.58                       & 0.44 & 0.56 & 0.60 & 0.58 & 0.46 & \textbf{0.59} & 0.57 & 0.50 \\
        \grayrow & \textbf{Mean} & 0.66 & 0.63 & 0.67 & 0.68 & 0.70 & 0.68 & \textbf{0.73}  & 0.64 & 0.70 & 0.70 & 0.71 & 0.65 & \textbf{0.73} & 0.71 & 0.59 \\

        \bottomrule[1.5pt]
        \multicolumn{17}{l}{$\bm{\bowtie}$ = fusion of modalities, Pup. = Pupil size, Int. = Pixel Intensity, F.f. = Fisherface features, $\mu$-E. = micro-expressions. } \\
        \end{tabular}}
        
    }
    \label{tab:continuous_affect_prediction_A_B_individual}
\end{table}
\addtolength{\tabcolsep}{0.3em}


\addtolength{\tabcolsep}{-0.3em}

\renewcommand{\arraystretch}{0.9}

\begin{table}[h]

    \centering
    \caption{\textbf{Standard deviation of continuous affect domain prediction results.} }
    \adjustbox{max width=\textwidth}{

{\small
        \begin{tabular}{p{1.8cm}p{1.7cm}cccc>{\hspace{1em}}ccccccccc>{\hspace{1em}}c>{\hspace{1em}}c}
        
        \toprule[1.5pt]
         &  & \multicolumn{4}{c}{\textbf{Wearable devices}} & \multicolumn{9}{c}{\textbf{Egocentric glasses}} & \textbf{All} & \textbf{Baseline} \\
         &  &  &  &  &  & \multicolumn{6}{c}{$\overbrace{\hspace{3.9cm}}^{\small{\textnormal{ET video}}}$} &  &  &  &  &  \\
        Session  & Domain & ECG & EDA & RSP & $\bm{\bowtie}$ & Pup. &  Int. & F.f. & Gaze & \hspace{-0.1em}B\hspace{-0.05em}l\hspace{-0.05em}i\hspace{-0.05em}n\hspace{-0.05em}k\hspace{-0.05em} & $\mu$-E. & PPG & IMU & $\bm{\bowtie}$ & $\bm{\bowtie}$ & Random \\
            \midrule
            \midrule
            \multirow[c]{3}{*}{\textbf{A}} & Arousal & 0.22 & 0.22 & 0.20 & 0.20 & 0.22 & 0.22 & 0.22   & 0.24 & 0.23 & 0.22 & 0.21 & 0.24 & 0.22 & 0.20 \\
             & Valence & 0.12 & 0.10 & 0.16 & 0.15 & 0.27 & 0.19 & 0.08                                 & 0.15 & 0.05 & 0.10 & 0.08 & 0.11 & 0.14 & 0.14 \\
             & Dominance & 0.23 & 0.24 & 0.24 & 0.24 & 0.22 & 0.23 & 0.22                               & 0.24 & 0.24 & 0.24 & 0.23 & 0.24 & 0.24 & 0.24 \\
             \grayrow & \textbf{Mean} & 0.19 & 0.19 & 0.20 & 0.20 & 0.24 & 0.21 & 0.17                  & 0.21 & 0.17 & 0.19 & 0.17 & 0.20 & 0.20 & 0.20 \\
            
             &  &  &  &  &  &  &  &  &  &  &  &   \\
            
            \multirow[c]{3}{*}{\textbf{B}} & Arousal & 0.16 & 0.16 & 0.16 & 0.16 & 0.16 & 0.16 & 0.16   & 0.16 & 0.16 & 0.16 & 0.16 & 0.16 & 0.16 & 0.16 \\
             & Valence & 0.16 & 0.16 & 0.16 & 0.15 & 0.14 & 0.13 & 0.14                                 & 0.16 & 0.19 & 0.18 & 0.15 & 0.18 & 0.14 & 0.14 \\
             & Dominance & 0.23 & 0.26 & 0.23 & 0.25 & 0.25 & 0.22 & 0.26                               & 0.25 & 0.23 & 0.26 & 0.22 & 0.22 & 0.26 & 0.22 \\
            \grayrow & \textbf{Mean} & 0.18 & 0.19 & 0.18 & 0.19 & 0.18 & 0.17 & 0.19                   & 0.19 & 0.19 & 0.20 & 0.18 & 0.19 & 0.19 & 0.17 \\
             &  &  &  &  &  &  &  &  &  &  &  &   \\
            
            \multirow[c]{3}{*}{\textbf{A+B}} & Arousal & 0.14 & 0.13 & 0.13 & 0.14 & 0.14 & 0.14 & 0.13 & 0.14 & 0.14 & 0.13 & 0.14 & 0.13 & 0.15 & 0.14 \\
             & Valence & 0.10 & 0.11 & 0.10 & 0.09 & 0.10 & 0.10 & 0.06                                 & 0.09 & 0.09 & 0.09 & 0.09 & 0.10 & 0.10 & 0.09 \\
             & Dominance & 0.18 & 0.18 & 0.17 & 0.18 & 0.17 & 0.19 & 0.16                               & 0.19 & 0.19 & 0.20 & 0.17 & 0.18 & 0.18 & 0.17 \\
             \grayrow & \textbf{Mean} & 0.14 & 0.14 & 0.13 & 0.14 & 0.14 & 0.14 & 0.12                  & 0.14 & 0.14 & 0.14 & 0.14 & 0.14 & 0.14 & 0.13 \\
            
            \bottomrule[1.5pt]
            \multicolumn{17}{l}{$\bm{\bowtie}$ = fusion of modalities, Pup. = Pupil size, Int. = Pixel Intensity, F.f. = Fisherface features, $\mu$-E. = micro-expressions. } \\
            \end{tabular}}
        
    }
    \vspace{2em}
    \label{tab:continuous_affect_prediction_STD}
\end{table}

\addtolength{\tabcolsep}{0.3em}

\subsection{Discrete Emotion Prediction}

Table~\ref{tab:detailed_emotion_prediction_A_B_individual} presents the discrete emotion prediction results in session A and session B. 
The egocentric glasses significantly exceeded the physiological sensors in predicting 9 discrete emotions ($F_1 =$~0.55 vs. $F_1 =$~0.25) in session A. 
In the naturalistic tasks, the egocentric glasses and the wearable devices had comparable results, with $F_1 =$~0.40 and $F_1 =$~0.33, respectively. With the current feature extraction, emotions such as sadness (0.87) and anger (0.68) had high prediction scores in session A using the combined sensors from the egocentric glasses.
The emotion of awe proved to be difficult to predict across experiments, with discrete emotions in session A achieving higher prediction results in comparison to session B.
Disgust has a high $F_1$ score in session B (0.75) when predicting it from the egocentric glasses.
The eye pupil size was highly informative for predicting fear in participants during session B ($F_1 =$~0.66).

Table~\ref{tab:std_detailed_emotion_prediction_A_B_individual} presents the standard deviations of the discrete emotion prediction results.
Session A tends to be noisier, with larger fluctuations in certain cases (e.g., IMU and F.f.), while Session B looks more consistent overall. 

\vspace{-1em}
\addtolength{\tabcolsep}{-0.3em}
\begin{table}[H]
    \centering
    \caption{\textbf{Discrete emotion prediction results.}}
    \adjustbox{max width=\textwidth}{


{\small
        \begin{tabular}{p{1.8cm}p{1.7cm}cccc>{\hspace{1em}}ccccccccc>{\hspace{1em}}c>{\hspace{1em}}c}
        
        \toprule[1.5pt]
         &  & \multicolumn{4}{c}{\textbf{Wearable devices}} & \multicolumn{9}{c}{\textbf{Egocentric glasses}} & \textbf{All} & \textbf{Baseline} \\
         &  &  &  &  &  & \multicolumn{6}{c}{$\overbrace{\hspace{3.9cm}}^{\small{\textnormal{ET video}}}$} &  &  &  &  &  \\
        Session  & Domain & ECG & EDA & RSP & $\bm{\bowtie}$ & Pup. &  Int. & F.f. & Gaze & \hspace{-0.1em}B\hspace{-0.05em}l\hspace{-0.05em}i\hspace{-0.05em}n\hspace{-0.05em}k\hspace{-0.05em} & $\mu$-E. & PPG & IMU & $\bm{\bowtie}$ & $\bm{\bowtie}$ & Random \\
        \midrule
        \midrule
        \multirow[c]{9}{*}{\textbf{A}} & Amused & 0.05 & 0.52 & 0.37 & 0.50 & 0.60 & 0.19 & \textbf{0.62}   & 0.47 & 0.17 & 0.17 & 0.52 & 0.54 & 0.52 & 0.57 & 0.12 \\
         & Content & 0.25 & 0.19 & 0.20 & 0.26 & 0.38 & 0.13 & 0.56                                         & 0.26 & 0.13 & 0.13 & 0.44 & 0.40 & \textbf{0.61} & \textbf{0.61} & 0.16 \\
         & Excited & 0.00 & 0.00 & 0.00 & 0.00 & 0.09 & 0.00 & 0.25                                         & 0.00 & 0.06 & 0.00 & 0.07 & 0.00 & \textbf{0.29} & 0.24 & 0.05 \\
         & Awe & 0.05 & 0.00 & 0.07 & 0.00 & \textbf{0.38} & 0.06 & 0.28                                    & 0.00 & 0.05 & 0.05 & 0.00 & 0.06 & 0.25 & 0.34 & 0.07 \\
         & Neutral & 0.23 & 0.36 & 0.31 & 0.32 & 0.40 & 0.24 & 0.49                                         & 0.44 & 0.27 & 0.32 & 0.37 & 0.41 & 0.51 & \textbf{0.52} & 0.21 \\
         & Fear & 0.00 & 0.24 & 0.00 & 0.06 & 0.48 & 0.08 & 0.52                                            & 0.11 & 0.00 & 0.00 & \textbf{0.60} & 0.12 & 0.57 & 0.53 & 0.06 \\
         & Sad & 0.25 & 0.82 & 0.23 & \textbf{0.88} & 0.77 & 0.07 & 0.85                                    & 0.68 & 0.13 & 0.13 & \textbf{0.88} & 0.75 & 0.87 & 0.87 & 0.10 \\
         & Disgust & 0.07 & 0.18 & 0.17 & 0.12 & 0.20 & 0.19 & 0.47                                         & 0.09 & 0.15 & 0.16 & 0.57 & 0.27 & \textbf{0.63} & 0.60 & 0.13 \\
         & Anger & 0.12 & 0.00 & 0.12 & 0.14 & 0.32 & 0.08 & 0.66                                           & 0.06 & 0.16 & 0.05 & 0.14 & 0.12 & \textbf{0.68} & \textbf{0.68} & 0.09 \\
         \grayrow & \textbf{Mean} & 0.11 & 0.26 & 0.16 & 0.25 & 0.40 & 0.12 & 0.52                          & 0.23 & 0.12 & 0.11 & 0.40 & 0.30 & \textbf{0.55} & \textbf{0.55} & 0.11 \\ 
         
         &  &  &  &  &  &  &  &  &  &  &  &  &  \\
         
        \multirow[c]{9}{*}{\textbf{B}} & Amused & 0.57 & 0.50 & 0.58 & 0.60 & 0.50 & 0.48 & \textbf{0.62}   & 0.44 & 0.44 & 0.51 & 0.58 & 0.52 & 0.61 & 0.60 & 0.32 \\
         & Content & 0.44 & 0.19 & 0.38 & 0.48 & 0.24 & 0.24 & \textbf{0.58}                                & 0.24 & 0.18 & 0.28 & 0.46 & 0.28 & 0.57 & 0.57 & 0.15 \\
         & Excited & 0.00 & 0.00 & 0.00 & 0.00 & 0.00 & 0.00 & 0.11                                         & 0.00 & 0.13 & 0.00 & 0.00 & 0.00 & \textbf{0.19} & 0.10 & 0.05 \\
         & Awe & 0.00 & 0.00 & 0.00 & 0.00 & 0.00 & 0.00 & 0.00                                             & 0.00 & 0.00 & 0.00 & 0.00 & 0.00 & 0.00 & 0.00 & \textbf{0.01} \\
         & Neutral & 0.04 & \textbf{0.25} & 0.13 & 0.05 & 0.10 & 0.14 & 0.00                                & 0.32 & 0.11 & 0.04 & 0.04 & 0.09 & 0.00 & 0.00 & 0.11 \\
         & Fear & 0.14 & 0.33 & 0.33 & 0.60 & 0.66 & 0.04 & 0.64                                            & 0.22 & 0.14 & 0.26 & 0.57 & 0.28 & \textbf{0.67} & 0.66 & 0.11 \\
         & Sad & 0.16 & 0.26 & 0.34 & 0.39 & 0.19 & 0.26 & \textbf{0.49}                                    & 0.47 & 0.20 & 0.29 & 0.37 & 0.05 & 0.44 & 0.42 & 0.10 \\
         & Disgust & 0.27 & 0.70 & 0.50 & 0.74 & 0.61 & 0.26 & 0.73                                         & 0.25 & 0.15 & 0.55 & 0.75 & 0.72 & 0.75 & \textbf{0.77} & 0.10 \\
         & Anger & 0.00 & 0.00 & 0.00 & 0.15 & 0.00 & 0.08 & 0.27                                           & 0.17 & 0.07 & 0.00 & 0.18 & 0.00 & \textbf{0.34} & 0.24 & 0.06 \\
        \grayrow & \textbf{Mean} & 0.18 & 0.25 & 0.25 & 0.33 & 0.26 & 0.17 & 0.38                           & 0.12 & 0.16 & 0.21 & 0.33 & 0.22 & \textbf{0.40} & 0.37 & 0.11 \\

        \bottomrule[1.5pt]
        \multicolumn{17}{l}{$\bm{\bowtie}$ = fusion of modalities, Pup. = Pupil size, Int. = Pixel Intensity, F.f. = Fisherface features, $\mu$-E. = micro-expressions. } \\
        \end{tabular}}

    }    
    \vspace{-3em}
    \label{tab:detailed_emotion_prediction_A_B_individual}
\end{table}
\addtolength{\tabcolsep}{0.3em}

\addtolength{\tabcolsep}{-0.3em}
\begin{table}[H]
    \centering
    \caption{\textbf{Standard deviation of discrete emotion prediction results.}}
    \adjustbox{max width=\textwidth}{


{\small
        \begin{tabular}{p{1.8cm}p{1.7cm}cccc>{\hspace{1em}}ccccccccc>{\hspace{1em}}c>{\hspace{1em}}c}
        
        \toprule[1.5pt]
         &  & \multicolumn{4}{c}{\textbf{Wearable devices}} & \multicolumn{9}{c}{\textbf{Egocentric glasses}} & \textbf{All} & \textbf{Baseline} \\
         &  &  &  &  &  & \multicolumn{6}{c}{$\overbrace{\hspace{3.9cm}}^{\small{\textnormal{ET video}}}$} &  &  &  &  &  \\
        Session  & Domain & ECG & EDA & RSP & $\bm{\bowtie}$ & Pup. &  Int. & F.f. & Gaze & \hspace{-0.1em}B\hspace{-0.05em}l\hspace{-0.05em}i\hspace{-0.05em}n\hspace{-0.05em}k\hspace{-0.05em} & $\mu$-E. & PPG & IMU & $\bm{\bowtie}$ & $\bm{\bowtie}$ & Random \\
        \midrule
        \midrule
        \multirow[c]{9}{*}{\textbf{A}} & Amused & 0.16 & 0.46 & 0.40 & 0.45 & 0.44 & 0.32 & 0.46    & 0.44 & 0.25 & 0.25 & 0.44 & 0.45 & 0.46 & 0.47 \\
         & Content & 0.26 & 0.23 & 0.24 & 0.31 & 0.32 & 0.25 & 0.35                                 & 0.24 & 0.15 & 0.20 & 0.39 & 0.32 & 0.34 & 0.35 \\
         & Excited & 0.00 & 0.00 & 0.00 & 0.00 & 0.10 & 0.00 & 0.28                                 & 0.00 & 0.11 & 0.00 & 0.16 & 0.00 & 0.28 & 0.28  \\
         & Awe & 0.16 & 0.00 & 0.10 & 0.00 & 0.41 & 0.10 & 0.34                                     & 0.00 & 0.06 & 0.11 & 0.00 & 0.16 & 0.36 & 0.42 \\
         & Neutral & 0.26 & 0.32 & 0.25 & 0.26 & 0.28 & 0.23 & 0.33                                 & 0.32 & 0.26 & 0.28 & 0.32 & 0.29 & 0.32 & 0.33 \\
         & Fear & 0.00 & 0.28 & 0.00 & 0.16 & 0.43 & 0.16 & 0.47                                    & 0.22 & 0.00 & 0.00 & 0.47 & 0.22 & 0.47 & 0.46 \\
         & Sad & 0.32 & 0.42 & 0.33 & 0.38 & 0.45 & 0.19 & 0.40                                     & 0.47 & 0.17 & 0.28 & 0.36 & 0.44 & 0.38 & 0.38 \\
         & Disgust & 0.16 & 0.32 & 0.30 & 0.21 & 0.29 & 0.31 & 0.45                                 & 0.25 & 0.21 & 0.28 & 0.47 & 0.38 & 0.44 & 0.44 \\
         & Anger & 0.24 & 0.00 & 0.24 & 0.30 & 0.38 & 0.19 & 0.48                                   & 0.17 & 0.25 & 0.16 & 0.28 & 0.24 & 0.46 & 0.46 \\
         \grayrow & \textbf{Mean} & 0.17 & 0.23 & 0.21 & 0.23 & 0.34 & 0.19 & 0.40                  & 0.11 & 0.08 & 0.11 & 0.32 & 0.28 & 0.39 & 0.40 \\
        
         &  &  &  &  &  &  &   &  &  &  &  &  \\
        
        \multirow[c]{9}{*}{\textbf{B}} & Amused & 0.25 & 0.28 & 0.29 & 0.28 & 0.25 & 0.27 & 0.28    & 0.26 & 0.28 & 0.28 & 0.25 & 0.22 & 0.31 & 0.30 \\
         & Content & 0.39 & 0.32 & 0.40 & 0.44 & 0.29 & 0.36 & 0.43                                 & 0.23 & 0.31 & 0.34 & 0.44 & 0.37 & 0.41 & 0.42  \\
         & Excited & 0.00 & 0.00 & 0.00 & 0.00 & 0.00 & 0.00 & 0.16                                 & 0.00 & 0.16 & 0.00 & 0.00 & 0.00 & 0.22 & 0.16  \\
         & Awe & 0.00 & 0.00 & 0.00 & 0.00 & 0.00 & 0.00 & 0.00                                     & 0.00 & 0.00 & 0.00 & 0.00 & 0.00 & 0.00 & 0.00 \\
         & Neutral & 0.00 & 0.31 & 0.18 & 0.10 & 0.11 & 0.21 & 0.00                                 & 0.00 & 0.16 & 0.11 & 0.08 & 0.13 & 0.00 & 0.00 \\
         & Fear & 0.30 & 0.39 & 0.36 & 0.49 & 0.49 & 0.08 & 0.50                                    & 0.46 & 0.28 & 0.30 & 0.48 & 0.36 & 0.49 & 0.49  \\
         & Sad & 0.28 & 0.33 & 0.41 & 0.42 & 0.26 & 0.33 & 0.46                                     & 0.34 & 0.32 & 0.37 & 0.41 & 0.10 & 0.43 & 0.43 \\
         & Disgust & 0.27 & 0.51 & 0.45 & 0.50 & 0.50 & 0.36 & 0.50                                 & 0.50 & 0.28 & 0.48 & 0.50 & 0.50 & 0.50 & 0.49 \\
         & Anger & 0.00 & 0.00 & 0.00 & 0.22 & 0.00 & 0.16 & 0.28                                   & 0.16 & 0.16 & 0.00 & 0.24 & 0.00 & 0.31 & 0.28 \\
         \grayrow & \textbf{Mean} & 0.17 & 0.24 & 0.23 & 0.27 & 0.21 & 0.20 & 0.29                  & 0.16 & 0.18 & 0.21 & 0.27 & 0.19 & 0.30 & 0.29 \\
        
         &  &  &  &  &  &  &  &  &  &  &  &  \\
        
        \multirow[c]{9}{*}{\textbf{A+B}} & Amused & 0.22 & 0.23 & 0.22 & 0.22 & 0.22 & 0.18 & 0.21  & 0.22 & 0.19 & 0.32 & 0.18 & 0.23 & 0.21 & 0.22 \\
         & Content & 0.23 & 0.21 & 0.26 & 0.24 & 0.26 & 0.23 & 0.30                                 & 0.22 & 0.24 & 0.20 & 0.28 & 0.22 & 0.24 & 0.24  \\
         & Excited & 0.00 & 0.10 & 0.00 & 0.00 & 0.19 & 0.00 & 0.18                                 & 0.00 & 0.00 & 0.00 & 0.00 & 0.00 & 0.20 & 0.18 \\
         & Awe & 0.10 & 0.00 & 0.00 & 0.00 & 0.32 & 0.00 & 0.34                                     & 0.16 & 0.08 & 0.00 & 0.00 & 0.00 & 0.43 & 0.40 \\
         & Neutral & 0.20 & 0.25 & 0.21 & 0.17 & 0.24 & 0.19 & 0.28                                 & 0.26 & 0.21 & 0.17 & 0.24 & 0.24 & 0.29 & 0.28 \\
         & Fear & 0.18 & 0.24 & 0.26 & 0.38 & 0.39 & 0.25 & 0.38                                    & 0.36 & 0.13 & 0.20 & 0.41 & 0.16 & 0.38 & 0.40 \\
         & Sad & 0.30 & 0.35 & 0.27 & 0.35 & 0.34 & 0.21 & 0.35                                     & 0.31 & 0.17 & 0.12 & 0.38 & 0.33 & 0.32 & 0.34 \\
         & Disgust & 0.16 & 0.28 & 0.27 & 0.32 & 0.33 & 0.25 & 0.35                                 & 0.29 & 0.19 & 0.20 & 0.36 & 0.34 & 0.34 & 0.32 \\
         & Anger & 0.00 & 0.18 & 0.22 & 0.19 & 0.30 & 0.00 & 0.42                                   & 0.21 & 0.19 & 0.03 & 0.31 & 0.27 & 0.44 & 0.40 \\
         \grayrow & \textbf{Mean} & 0.15 & 0.20 & 0.19 & 0.21 & 0.29 & 0.15 & 0.31                  & 0.08 & 0.05 & 0.14 & 0.24 & 0.20 & 0.32 & 0.31 \\
        
        \bottomrule[1.5pt]
        \multicolumn{17}{l}{$\bm{\bowtie}$ = fusion of modalities, Pup. = Pupil size, Int. = Pixel Intensity, F.f. = Fisherface features, $\mu$-E. = micro-expressions. } \\
        \end{tabular}}

    }
    \label{tab:std_detailed_emotion_prediction_A_B_individual}
\end{table}
\addtolength{\tabcolsep}{0.3em}

\vspace{-1em}

\subsection{Personality Prediction}

Table~\ref{tab:personality_prediction_A_B_individual} presents the personality trait predictions for session A and session B.
The pupil size and blink rate achieved the highest $F_1$ score in session A, while ECG performed best in session B ($F_1 =$~0.60).
Table~\ref{tab:personality_prediction_STD} presents the standard deviations of the personality prediction results. 

\addtolength{\tabcolsep}{-0.3em}
\begin{table}[h]
    \centering
    \caption{\textbf{Personality prediction results.}}
    \adjustbox{max width=\textwidth}{


{\small
        \begin{tabular}{p{1.8cm}p{1.7cm}cccc>{\hspace{1em}}ccccccccc>{\hspace{1em}}c>{\hspace{1em}}c}
        
        \toprule[1.5pt]
         &  & \multicolumn{4}{c}{\textbf{Wearable devices}} & \multicolumn{9}{c}{\textbf{Egocentric glasses}} & \textbf{All} & \textbf{Baseline} \\
         &  &  &  &  &  & \multicolumn{6}{c}{$\overbrace{\hspace{3.9cm}}^{\small{\textnormal{ET video}}}$} &  &  &  &  &  \\
        Session  & Domain & ECG & EDA & RSP & $\bm{\bowtie}$ & Pup. &  Int. & F.f. & Gaze & \hspace{-0.1em}B\hspace{-0.05em}l\hspace{-0.05em}i\hspace{-0.05em}n\hspace{-0.05em}k\hspace{-0.05em} & $\mu$-E. & PPG & IMU & $\bm{\bowtie}$ & $\bm{\bowtie}$ & Random \\
        \midrule
        \midrule
        \multirow[c]{5}{*}{\textbf{A}} & Ex & 0.48 & 0.48 & 0.42 & 0.30 & 0.48 & 0.48 & 0.45    & 0.48 & \textbf{0.58} & 0.38 & 0.45 & 0.48 & 0.38 & 0.25 & \textbf{0.55} \\
         & Ag & 0.45 & 0.42 & 0.48 & 0.38 & 0.42 & 0.32 & 0.22                                  & 0.40 & \textbf{0.58} & 0.45 & 0.57 & 0.50 & 0.48 & 0.52 & 0.52 \\
         & Co & 0.42 & 0.52 & 0.28 & 0.42 & \textbf{0.65} & 0.48 & 0.52                         & 0.58 & 0.63 & 0.63 & 0.52 & 0.40 & 0.35 & 0.38 & 0.55 \\
         & NE & 0.50 & 0.52 & 0.50 & 0.52 & \textbf{0.68} & 0.45 & 0.30                         & 0.58 & 0.45 & 0.53 & 0.50 & 0.57 & 0.62 & 0.48 & 0.52 \\
         & OM & 0.30 & 0.60 & \textbf{0.62} & 0.48 & 0.52 & 0.55 & 0.42                         & 0.60 & 0.53 & 0.45 & 0.28 & 0.60 & 0.60 & 0.45 & 0.52 \\
        \grayrow & \textbf{Mean} & 0.43 & 0.51 & 0.46 & 0.42 & \textbf{0.55} & 0.46 & 0.38      & 0.53 & \textbf{0.55} & 0.49 & 0.46 & 0.51 & 0.49 & 0.42 & 0.53\\
        
         &  &  &  &  &  &  &  &  &  &  &  & & & &  &  \\
        
        \multirow[c]{5}{*}{\textbf{B}} & Ex & 0.45 & 0.50 & 0.48 & 0.55 & 0.45 & \textbf{0.75} & 0.53 & 0.45 & 0.73 & 0.52 & 0.55 & \textbf{0.75} & 0.48 & 0.55 \\
         & Ag & 0.48 & 0.60 & 0.48 & 0.48 & 0.48 & 0.60 & 0.45                                 & \textbf{0.70} & 0.45 & 0.63 & 0.38 & 0.32 & 0.65 & 0.35 & 0.52 \\
         & Co & 0.68 & 0.40 & 0.35 & 0.62 & 0.45 & 0.28 & 0.40                         &\textbf{0.70} & 0.40 & 0.45 & 0.62 & 0.50 & 0.45 & 0.62 & 0.55 \\
         & NE & \textbf{0.78} & 0.52 & 0.65 & 0.70 & 0.60 & 0.52 & 0.42                         & 0.53 & 0.43 & 0.73 & 0.48 & 0.65 & 0.40 & 0.72 & 0.52 \\
         & OM & \textbf{0.62} & 0.52 & 0.57 & 0.35 & 0.48 & 0.50 & 0.40                         & 0.48 & 0.53 & 0.43 & 0.52 & 0.40 & 0.35 & 0.28 & 0.52 \\
         \grayrow & \textbf{Mean} & \textbf{0.60} & 0.51 & 0.51 & 0.54 & 0.49 & 0.53 & 0.44     & 0.59 & 0.45 & 0.59 & 0.50 & 0.48 & 0.52 & 0.49 & 0.53\\
            
        \bottomrule[1.5pt]
        \multicolumn{17}{l}{$\bm{\bowtie}$ = fusion of modalities, Pup. = Pupil size, Int. = Pixel Intensity, F.f. = Fisherface features, $\mu$-E. = micro-expressions. } \\
        \end{tabular}}

    }
    \label{tab:personality_prediction_A_B_individual}
\end{table}
\addtolength{\tabcolsep}{0.3em}

\addtolength{\tabcolsep}{-0.3em}
\begin{table}[h]
    \centering
    \caption{\textbf{Standard deviation of personality prediction results.}}
    \adjustbox{max width=\textwidth}{

    {\small
        \begin{tabular}{p{1.8cm}p{1.7cm}cccc>{\hspace{1em}}ccccccccc>{\hspace{1em}}c>{\hspace{1em}}c}
        
        \toprule[1.5pt]
         &  & \multicolumn{4}{c}{\textbf{Wearable devices}} & \multicolumn{9}{c}{\textbf{Egocentric glasses}} & \textbf{All} & \textbf{Baseline} \\
         &  &  &  &  &  & \multicolumn{6}{c}{$\overbrace{\hspace{3.9cm}}^{\small{\textnormal{ET video}}}$} &  &  &  &  &  \\
        Session  & Domain & ECG & EDA & RSP & $\bm{\bowtie}$ & Pup. &  Int. & F.f. & Gaze & \hspace{-0.1em}B\hspace{-0.05em}l\hspace{-0.05em}i\hspace{-0.05em}n\hspace{-0.05em}k\hspace{-0.05em} & $\mu$-E. & PPG & IMU & $\bm{\bowtie}$ & $\bm{\bowtie}$ & Random \\
        \midrule
        \midrule
        \multirow[c]{5}{*}{\textbf{A}} & Ex & 0.50 & 0.50 & 0.49 & 0.48 & 0.50 & 0.50 & 0.50 & 0.50 & 0.49 & 0.48 & 0.50 & 0.50 & 0.48 & 0.45 \\
         & Ag & 0.50 & 0.49 & 0.50 & 0.50 & 0.49 & 0.47 & 0.42 & 0.46 & 0.49 & 0.50 & 0.49 & 0.50 & 0.50 & 0.50 \\
         & Co & 0.49 & 0.50 & 0.45 & 0.50 & 0.48 & 0.50 & 0.50 & 0.46 & 0.48 & 0.48 & 0.50 & 0.49 & 0.48 & 0.48 \\
         & NE & 0.50 & 0.50 & 0.50 & 0.50 & 0.47 & 0.50 & 0.46 & 0.50 & 0.50 & 0.50 & 0.50 & 0.49 & 0.48 & 0.50 \\
         & OM & 0.46 & 0.49 & 0.48 & 0.50 & 0.50 & 0.50 & 0.49 & 0.50 & 0.50 & 0.50 & 0.45 & 0.49 & 0.49 & 0.50 \\
         \grayrow & \textbf{Mean} & 0.49 & 0.50 & 0.48 & 0.50 & 0.49 & 0.50 & 0.47 & 0.48 & 0.49 & 0.49 & 0.49 & 0.49 & 0.49 & 0.49 \\
        
         &  &  &  &  &  &  &  &  &  &  &  &  \\
        
        \multirow[c]{5}{*}{\textbf{B}} & Ex & 0.49 & 0.50 & 0.50 & 0.50 & 0.50 & 0.43 & 0.50 & 0.49 & 0.50 & 0.45 & 0.50 & 0.50 & 0.43 & 0.50 \\
         & Ag & 0.50 & 0.49 & 0.50 & 0.50 & 0.50 & 0.49 & 0.50 & 0.49 & 0.50 & 0.48 & 0.48 & 0.47 & 0.48 & 0.47 \\
         & Co & 0.48 & 0.49 & 0.48 & 0.48 & 0.50 & 0.45 & 0.49 & 0.50 & 0.49 & 0.50 & 0.48 & 0.50 & 0.50 & 0.48 \\
         & NE & 0.47 & 0.50 & 0.48 & 0.46 & 0.49 & 0.50 & 0.49 & 0.49 & 0.49 & 0.45 & 0.50 & 0.48 & 0.49 & 0.47 \\
         & OM & 0.49 & 0.50 & 0.49 & 0.48 & 0.50 & 0.50 & 0.49 & 0.50 & 0.50 & 0.49 & 0.50 & 0.49 & 0.48 & 0.45 \\
         \grayrow & \textbf{Mean} & 0.49 & 0.50 & 0.49 & 0.48 & 0.50 & 0.47 & 0.49 & 0.50 & 0.50 & 0.47 & 0.49 & 0.49 & 0.48 & 0.47 \\
        
         &  &  &  &  &  &  &  &  &  &  &  &  \\
        
        \multirow[c]{5}{*}{\textbf{A+B}} & Ex & 0.43 & 0.50 & 0.47 & 0.40 & 0.49 & 0.46 & 0.50 & 0.49 & 0.49 & 0.50 & 0.49 & 0.50 & 0.50 & 0.50 \\
         & Ag & 0.48 & 0.49 & 0.50 & 0.49 & 0.50 & 0.49 & 0.49 & 0.49 & 0.48 & 0.49 & 0.49 & 0.49 & 0.46 & 0.49 \\
         & Co & 0.49 & 0.50 & 0.46 & 0.50 & 0.50 & 0.49 & 0.49 & 0.50 & 0.49 & 0.50 & 0.48 & 0.50 & 0.50 & 0.48 \\
         & NE & 0.50 & 0.50 & 0.48 & 0.47 & 0.47 & 0.49 & 0.46 & 0.49 & 0.50 & 0.49 & 0.49 & 0.50 & 0.48 & 0.45 \\
         & OM & 0.48 & 0.50 & 0.50 & 0.49 & 0.50 & 0.49 & 0.49 & 0.50 & 0.48 & 0.47 & 0.46 & 0.48 & 0.45 & 0.48\\
        \grayrow & \textbf{Mean} & 0.48 & 0.50 & 0.48 & 0.47 & 0.49 & 0.48 & 0.49 & 0.50 & 0.49 & 0.49 & 0.48 & 0.49 & 0.48 & 0.48 \\
        
        \bottomrule[1.5pt]
        \multicolumn{17}{l}{$\bm{\bowtie}$ = fusion of modalities, Pup. = Pupil size, Int. = Pixel Intensity, F.f. = Fisherface features, $\mu$-E. = micro-expressions. } \\
        \end{tabular}}

}
\label{tab:personality_prediction_STD}
\end{table}
\addtolength{\tabcolsep}{0.3em}

\addtolength{\tabcolsep}{-0.2em}
\begin{table}[t]
\centering
\caption{\textbf{Detailed description of the emotion-inducing video clips.}}
{\small
\begin{tabular}{p{0.3cm}p{2.5cm}p{1.5cm}p{7.5cm}p{1.2cm}}
\toprule[1.5pt]
\textbf{ID} & \textbf{Video Label} & \textbf{Target Emotion} & \textbf{Description}  & \textbf{Duration (s)}\\
\midrule
\midrule
\texttt{1} & AnimalCruelty & Anger & Televised news of a dog groomer abusing dogs.	 & 40 \\
\texttt{2} & AuroraBorealis & Awe & A timelapse of the northern lights. & 40 \\
\texttt{3} & BearGrylls & Disgust & A man eats a worm. & 40 \\
\texttt{4} & CollegeAcceptance & Excitement & A student gets accepted to his dream college. & 40 \\
\texttt{5} & HarrySally & Amusement & Sally shows Harry how women fake orgasms at a restaurant. & 72 \\
\texttt{6} & JojoRabbit & Sadness & A boy embraces his mother who has been hanged. & 46 \\
\texttt{7} & LoveActually & Content & Narrator purporting that "love is everywhere". & 42 \\
\texttt{8} & MovingShapes & Neutral & Shapes moving on a neutral background. & 40 \\
\texttt{9} & Psycho & Fear & A lady gets murdered in her bathtub by an intruder. & 45 \\
\bottomrule[1.5pt]
\end{tabular}}
\label{tab:description_video_clips}
\vspace{-1em}
\end{table}
\addtolength{\tabcolsep}{0.2em}

\end{document}